\documentclass[runningheads]{llncs}

\usepackage{eccv}

\usepackage{eccvabbrv}

\usepackage{graphicx}
\usepackage{booktabs}

\usepackage[table]{xcolor}
\usepackage{wrapfig}

\usepackage{arydshln}
\usepackage{pifont}
\usepackage{multirow}
\usepackage{tabularx}
\usepackage{makecell}
\usepackage{caption}

\usepackage{amsmath, amssymb}

\usepackage{xcolor}

\usepackage[ruled]{algorithm2e}

\SetKwComment{Comment}{$\triangleright$ }{}

\SetCommentSty{algocommentsty}

\definecolor{MyBlue}{HTML}{e7f5ff}
\definecolor{MyTextBlue}{HTML}{364fc7}
\definecolor{MyTextRed}{HTML}{5f3dc4}
\DeclareMathAlphabet{\mathbbold}{U}{bbold}{m}{n}

\newcommand{\MODELNAME}{\textsc{Qwerty}\xspace}
\newcommand{\myxmark}{\ding{55}}
\newcommand{\myparagraph}[1]{\vspace{2pt}\noindent{\bf #1}}

\usepackage[accsupp]{axessibility}  % Improves PDF readability for those with disabilities.

\usepackage{tocloft}

\usepackage[breaklinks,colorlinks,citecolor=eccvblue]{hyperref}
\usepackage{orcidlink}

\begin{document}

\title{QWERTY: Training-Free Motion Control via Query-Warped Video Diffusion Transformers}

\titlerunning{Training-Free Motion Control via Query-Warped Video DiTs}

\author{
Kyobin Choo\inst{1}\orcidlink{0000-0003-2856-402X} \and
Youngmin Kim\inst{2} \and
Hyunkyung Han\inst{2} \and
Geunrip Park\inst{2} \and
Chanyoung Kim\inst{2} \and
Sunyoung Jung\inst{2} \and
Seong Jae Hwang\inst{2}\thanks{Corresponding author}
}

\authorrunning{K. Choo et al.}

\institute{
Department of Computer Science, Yonsei University, Seoul, Republic of Korea \and
Department of Artificial Intelligence, Yonsei University, Seoul, Republic of Korea \\
\email{\{chu, winston1214, hhk, gunlip1210, chanyoung, sunyoungj, seongjae\} @yonsei.ac.kr}}

% \maketitle
\begingroup
\renewcommand{\addcontentsline}[3]{} % prevent maketitle/authcount from entering toc
\maketitle
\endgroup

\let\savedaddcontentsline\addcontentsline
\renewcommand{\addcontentsline}[3]{}

\begin{center}
    \centering
    \captionsetup{type=figure}
    \includegraphics[width=\linewidth]{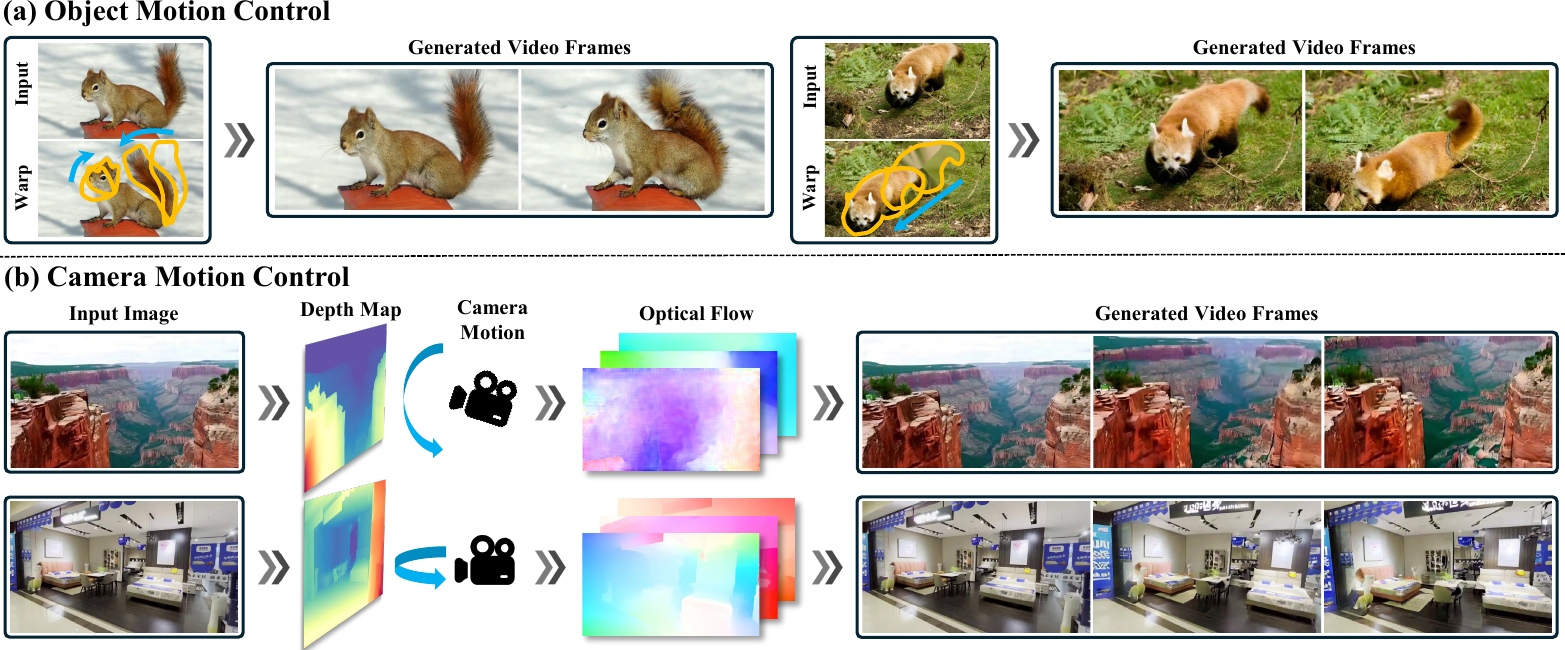}
    % \vspace{-20pt}
    \caption{
    We present \MODELNAME, a \textit{training-free} framework that controls the motion of an image-to-video diffusion transformer (DiT) according to user-defined warping. By warping the queries of the DiT at inference time, \MODELNAME steers motion while preserving the generative fidelity of the pretrained backbone. Our framework enables (a) fine-grained object control by translating, rotating, and scaling masks drawn on the input image, and (b) supports camera motion control by conditioning on optical flow.
    % We present \MODELNAME, 자유 형태의 user-defined warping에 따라 image-to-video diffusion transformer의 모션을 control하는 training-free framework. DiT의 query를 warping함으로써 ~~~. 우리의 방법론은 입력 이미지 위에서 직접 정의한 mask를 이동 회전 scale 시켜 fine한 object 컨트롤이 가능하고, optical flow를 입력받아 카메라 모션을 제어할 수 있다.
    }
    \label{fig:fig1}
\end{center}
% \twocolumn[{%
%     \renewcommand\twocolumn[1][]{#1}%
%     \maketitle
%     \input{sec/figs/fig_1}
% }]

\begin{abstract}

% Recent advances in video Diffusion Transformers (DiTs) have improved generation quality in fidelity, temporal coherence, and dynamics. Yet with text alone, users still resort to extensive prompt engineering and repeated resampling to realize desired motion.

Video diffusion transformers (DiTs) generate high-fidelity and temporally coherent videos, yet motion control remains implicit, primarily relying on text prompts. As a result, achieving desired motion often requires extensive prompt engineering and repeated resampling. While fine-tuning models with additional spatial prompts (\eg, bounding boxes or point trajectories) enables explicit control, it demands substantial data curation and computation, and may compromise the generative capabilities of pretrained models. Consequently, training-free motion control using such spatial prompts has been explored in U-Net–based video diffusion models, but remains largely unexplored for DiTs.
We introduce \MODELNAME, a training-free framework that enables flexible motion control in pretrained image-to-video DiTs via user-defined object warping and optical flow. We carefully manipulate the 3D full attention of DiTs by warping the frame-invariant semantic subspace of queries. We find that the noise predicted by the query-warped DiT naturally guides the diffusion trajectory toward the desired motion, and further show that leveraging this noise as self-guidance for latent optimization improves control stability and visual quality.
Experiments show that \MODELNAME achieves the most effective motion control among existing training-free approaches on a recent image-to-video DiT, with performance comparable to fine-tuning–based methods.
% Our code will be released upon acceptance.
\end{abstract}    
\section{Introduction}
\label{sec:intro}

% 1. Video model, motion control 필요성, 학습 기반 모델의 한계
Recent advances in video diffusion transformers (DiTs) enable the generation of high-quality videos with longer, more dynamic, and natural motion \cite{sora, veo3}. However, current pretrained models still control motion primarily through text prompts (\eg, \textit{make the squirrel wag its tail}), which are fundamentally ill-suited to precisely specifying where, how far, and how fast things should move \cite{motionprompting,sgi2v}. This mismatch has driven growing demand for more explicit motion control.
 
In response, many methods specify motion via spatial prompts such as masks \cite{animate_anything,motion-i2v}, bounding boxes \cite{magicmotion,sgi2v,moft}, and point trajectories \cite{motionpro}, and more recent approaches introduce richer representations like landmarks \cite{mofa_video}, free-form warping fields \cite{gowiththeflow} and multiple fine-grained trajectories \cite{motionprompting}. To inject such structural cues into pretrained models, existing methods typically add ControlNet-style \cite{controlnet} branches \cite{motionprompting}, warp the input noise \cite{gowiththeflow}, or encode the cues as extra tokens \cite{tora} and finetune the model. However, these approaches require resource-intensive training for each new model and can even degrade the generative capability of the pretrained backbone, which substantially undermines their practical usability.
% \cy{training-free가 필요한 이유가 이 motivation밖에 없나? 너무 당연한 이야기라 좀 킥이 없는듯. \eg, CASS에서는 training-free가 진정한 OVSS를 완성하는 길이다, train을 하는 순간 OVSS가 가져야 할 특유의 generalization capability가 없어질 수 있기 때문에 라고 함.}
% 최근 video Diffusion Transformers (DiTs)의 발전으로 더 길고 dynamic하고 자연스러운 모션을 가진 고품질 비디오를 생성할 수 있게 되었다. 그러나 최근 나온 pretrained 비디오 생성 모델도 여전히 대부분 텍스트로만 비디오의 모션을 표현할 수 밖에 없으며, 텍스트는 근본적으로 움직임의 위치, 크기, 속도 등을 명확하게 표현할 수 없기 때문에 보다 구체적인 motion control에 대한 요구가 커지고 있다. 모션을 명시적으로 표현하기 위해 일반적으로 mask, bounding box, point trjectory가 많이 사용되며, 최근에는 자유 형태의 warping이나 다중의 fine한 point trajectory 등 더 유연하고 구체적인 표현이 가능한 공간적 prompt가 정의되고 있다. 이러한 structural cue를 pretrained 모델에 주입하기 위해 일반적으로 ControlNet을 활용하거나, 입력 noise를 warping하거나, token 형태로 주입하여 모델을 finetuning한다. 그러나, 이는 방대한 양의 자원과 데이터가 필요할 뿐만 아니라, video 모델이 빠르게 발전하고 있는 지금, 새로운 모델이 나올 때마다 학습이 필요하다는 명확한 한계가 있다.

\begin{figure*}[t!]
    \begin{center}
        \includegraphics[width=\textwidth]{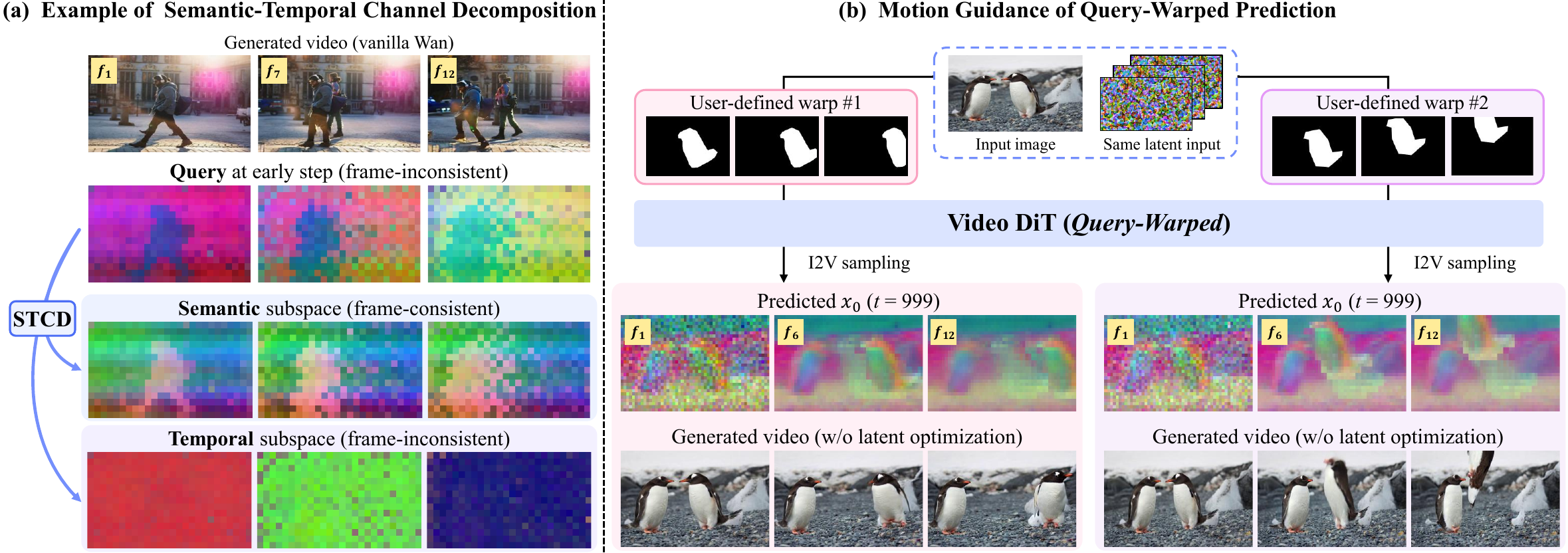}
    \end{center}
    \vspace{-12pt}
    \caption{Illustration of the two key components in our \textit{training-free} video motion control pipeline. {(a)} Our channel decomposition technique disentangles frame-inconsistent DiT query features into semantic and temporal subspace. The filtered semantic channels are frame-consistent and spatially discriminative, making them amenable to explicit motion control. {(b)} We warp queries to manipulate the 3D full attention of a video DiT, thereby steering motion generation. Even at an early timestep ($t{=}999$), starting from the same input latent, the query-warped DiT already provides distinct motion guidance that follows each user-specified warp, without any additional optimization.
    % \caption{Illustration of the two key components in our \textit{training-free} video motion control pipeline. \textbf{(a)} Our channel decomposition technique disentangles frame-inconsistent DiT query features into a semantic subspace and a temporal subspace. The filtered semantic channels are frame-consistent and spatially discriminative, making them easy to manipulate for motion control. \textbf{(b)} We warp queries in the Video DiT to effectively steer motion. Even at an early timestep ($t{=}999$), starting from the same input latent, the query-warped DiT already provides distinct motion guidance that follows each user-specified warp, without any additional optimization.
    % \caption{our proposed training-free video motion control pipeline의 두 핵심 component에 대한 효용성을 보여주는 그림. \textbf{(a)} 우리의 channel decomposition 방법은 프레임-일관적이지 않은 DiT query feature의 성분을 semantic subspace와 temporal subspace로 disentangle한다. 필터링된 semantic channel은 프레임 across로 일관적이고 공간-분별적이어서 조작이 용이하다. \textbf{(b)} 우리는 Video DiT 모델의 inference동안 query를 warp해서 원하는 모션을 효과적으로 유도한다. in early timestep(t=999), 같은 입력 latent에서 샘플링을 시작했음에도 모델이 예측한 noise가 각 warping에 알맞는 서로 다른 모션을 별도 optimize 없이도 가이드한 것을 볼 수 있다.
    }
    % \vspace{-9pt}
    \label{fig:fig_2}
\end{figure*}

% 2. training-free 방법에 대한 얘기: DiT는 아직 없음. Unet에서 잘 working하는 방법이 있으나 단점이 있고 DiT에 적합하지 않음.\
In light of these limitations, prior work on U-Net–based video diffusion models has explored training-free motion control \cite{3dtrajmaster,camtrol,draganything,imzero,anyi2v,smm}. Promising strategies developed for U-Nets—such as noise warping \cite{freetraj}, attention masking \cite{peekaboo}, and latent optimization based on bounding box similarity \cite{sgi2v, moft}—are either incompatible with the 3D full attention of DiTs or fail to function reliably in practice (Sec.~\ref{sec:experiments}).
More recent training-free motion transfer methods built on DiTs mimic motion from a reference video \cite{ropecraft, ditflow, DeT}. However, many practical scenarios require users to specify motion directly rather than relying on reference videos, highlighting the need for flexible, user-defined motion control in DiTs.
In this work, we introduce \MODELNAME, a training-free framework that enables user-defined control of object and camera motion in image-to-video (I2V) DiTs using mask warping or optical flow. Consistent with prior findings that temporal correspondences in DiTs are more salient in attention maps than in hidden states \cite{difftrack, ditflow}, our method steers motion generation by directly manipulating attention. As shown in the second row of Fig.~\ref{fig:fig_2}(a), the first-frame features of an I2V DiT exhibit a clear spatial layout even at early denoising steps, making them a reliable \textit{reference} feature map. We therefore warp the regions of first-frame queries corresponding to the desired motion and paste them into later frames, encouraging high attention between query tokens at frame $n$ and their matched key tokens from the first frame. Through this query-warping mechanism, we induce the desired correspondences within the 3D full attention, and demonstrate theoretically (Sec.~\ref{sec:why_query}) and empirically (Sec.~\ref{sec:ablation}) that warping queries is the only effective mechanism for achieving this behavior among all attention components.

However, consistent with observations in prior work \cite{DeT, sgi2v}, we find that DiT features are not frame-consistent (see the frame-variant query colors in Fig.~\ref{fig:fig_2}(a)). These frame-inconsistent components emerge after applying temporal rotary positional embeddings (RoPE)~\cite{rope}, and appear to encode the temporal ordering of frames rather than scene-level semantics. Consequently, pasting such queries across frames can distort the attention distribution and introduce visual artifacts.
To address this issue, we propose semantic–temporal channel decomposition (STCD). STCD computes channel-wise saliency scores by performing principal component analysis (PCA) separately on semantic and temporal anchor tokens, and then partitions the query channels into two groups based on these scores (Sec.~\ref{sec:channel_selection}). As shown in Fig.~\ref{fig:fig_2}(a), this decomposition yields a frame-consistent semantic subspace and a frame-variant temporal subspace within the query. We apply query warping only to the former, enabling controlled motion generation while minimizing distortion of the attention distribution (Sec.~\ref{sec:query_warping}).

% 4. QWERTY 아이디어 동기, query warping 디테일
% We verify that noise produced by a query-warped DiT 조정하다 diffusion path를 toward the intended motion. As shown in Fig.~\ref{fig:fig_2}(b), at the very first denoising step ($t=999$), the predicted $\mathbf{x}_0$ from the query-warped noise already exhibits the desired layout according to the user-defined warp. Starting from the same latent, different warps yield videos with corresponding motions, indicating that query warping provides strong motion guidance in the early steps (global motion이 형성되는~\cite{motionshop}). While query warping alone can guide motion, we further improve control stability and visual quality by using query-warped noise로 denoising process를 진행하는 것과 input latent를 optimize하는 것을 교차병행하며 (Sec.~\ref{sec:motion_guidance}). By harmoniously integrating our proposed components, \MODELNAME reveals an implicit motion prior encoded in DiT의 3D attention maps, which 부상할 수 있는 a key ingredient for practical training-free motion control.
We verify that the noise predicted by a query-warped DiT steers the diffusion path toward the intended motion. As shown in Fig.~\ref{fig:fig_2}(b), at the first denoising step ($t=999$), the predicted $\mathbf{x}_0$ derived from the query-warped noise already exhibits the desired layout according to the user-defined warp. Starting from the same latent, different warp configurations yield videos with motions consistent with the applied warps, indicating that query warping provides strong motion guidance in the early denoising steps, where global motion patterns form \cite{motionshop}. While this query-warped denoising alone can guide motion, we further improve control stability and visual quality by optimizing the input latent using query-warped noise as guidance (Sec.~\ref{sec:motion_guidance}). By integrating these components, we reveal an implicit motion prior encoded in the 3D full attention maps of DiTs, which may serve as a key ingredient for practical training-free motion control.

% \cy{정리하는 글 필요. \eg, By harmoniously integrating our proposed methods, our model captures ``an OOO attribute ", which is a key factor that must be considered in DiT training-free motion control.}

% 우리는 이렇게 the noise produced by query-warped DiT가 의도한 motion이 나타나도록 diffusion 샘플링을 가이드한다는 것을 확인하였다. As shown in Fig.~\ref{fig:fig_2}(b), at the very first denoising step ($t=999$), the predicted $\mathbf{x}_0$ computed from the query-warped noise already exhibits the desired layout according to the user-defined 와핑. 실제로 같은 latent noise에서 시작한 샘플링이 두 warping 입력에 맞는 motion을 가진 비디오의 생성으로 귀결되었으며, 이는 indicates that query warping at the earliest stages, where global motion is formed \cite{}, provides a strong motion guidance. Although 쿼리를 와핑하는것만으로 motion guide가 가능하지만, we further improve control stability and visual quality by using the query-warped noise both to run(다음스텝으로 넘어가는 동사?) the diffusion steps and to guide an optimization of the input latent (Sec.~\ref{sec:motion_guidance}).

% 4. contributions
\myparagraph{Contributions.} Our main contributions are as follows:
\begin{itemize}
% \item{We present \MODELNAME, the first training-free motion control framework specifically designed for image-to-video DiTs, supporting mask-warping–based object control and optical-flow–based camera control.}
\item{We present \MODELNAME, the first training-free motion control framework specifically designed for image-to-video DiTs, supporting flexible user-defined motion control via mask warping for objects and optical flow for cameras.}
\item{We show that our query warping enables \textit{explicit} manipulation of the 3D full attention in DiTs, and that the query-warped noise not only steers the diffusion path toward desired motion, but also serves as guidance for latent optimization, establishing a new mechanism for motion control in DiTs.}
% \item{We show that our query warping enables \textit{explicit} 조작 of the 3D full attention in DiTs, and that the resulting noise serves not only as a strong motion prior in its own right, but also as guidance for latent optimization. Together, these properties establish a new mechanism for motion control in DiTs.}
\item{We propose semantic–temporal channel decomposition (STCD), which enables motion control via query warping with minimal distortion of the attention distribution, thereby preserving visual quality. The anchoring mechanism is generally applicable for refining frame-inconsistent video features.}
\item{Extensive experiments show that \MODELNAME achieves state-of-the-art performance among training-free methods for object and camera motion control on DiTs, substantially narrowing the gap to fine-tuning–based approaches.}

% \item{우리는 user의 free-form 기반의 object motion, optical flow를 입력받아 camera motion을 control할 수 있는 최초의 training-free 범용, DiT전용 image-to-video framework인 \MODELNAME을 제안한다.}
% \item{query를 warping만으로 attention을 조작해 원하는 모션을 유도할 수 있는 기술을 보여줌}
% \item{우리는 ~Semantic-Temporal Channel Decomposition (STCD)를 제안하여 ~.}
% \item{우리는 ~Semantic-Temporal Channel Decomposition (STCD)를 제안한다.}
% \item{우리는 기성 및 자체 dataset을 통한 실험으로 object 그리고 camera motion control을 성공적으로 수행하고, 데이터셋train 기반의 DiT motion control 방법과의 격차를 크게 좁힌다.}
\end{itemize}

\section{Related Works}
\label{sec:related_works}

% Early image-to-video (I2V) diffusion models~\cite{videocrafter1} extended text-to-image (T2I) diffusion architectures~\cite{animatediff} based on UNet, generating video sequences from a single input image. 
% \ym{Version 1.}
% Early image-to-video (I2V) diffusion models~\cite{SVD, videocrafter1} predominantly adopted UNet architectures originally developed for text-to-image diffusion~\cite{animatediff}, expanding them to generate temporally extended sequences from a single image.
% However, UNet-based models inherently rely on CNN-driven spatial inductive biases, which limit temporal dependency modeling and the flexibility of input conditioning. Recent efforts have increasingly converged toward Diffusion Transformer~\cite{dit} (DiT) formulations, where spatio-temporal attention is modeled natively and text conditioning becomes an integral part of the generative process. Multiple recent studies report that DiT-based I2V approaches provide advantages in temporal smoothness, long-range dependency modeling, and overall video coherence.
% \ym{Version 2. }
% \myparagraph{Video Diffusion Models.}
\subsubsection{Video Diffusion Models.}
Following the success of text-to-image diffusion~\cite{ldm,dalle2,imagen}, early studies extended diffusion to video generation using U-Net-based architectures~\cite{videocrafter1,SVD,dynamicrafter}. These methods typically combine spatial attention within frames and temporal attention across tokens at the same spatial location to model video dynamics. However, this factorized design prevents direct interaction between tokens at different spatial locations across frames within a single operation, limiting the modeling of globally coherent motion. More recently, diffusion transformer (DiT)~\cite{dit} based video models employ 3D full attention to jointly model all tokens across space and time, enabling stronger temporal coherence~\cite{cogvideox,wan,hunyuanvideo,latte}. Motivated by this architectural shift toward DiT backbones, our query-warping mechanism leverages the 3D full-attention structure of DiTs.
\subsubsection{Motion-controllable Video Generation.}
To enhance the practical usability of video generation, it is important to directly control desired motion rather than merely producing plausible dynamics. Early approaches relied on intuitive spatial prompts such as bounding boxes~\cite{imzero,peekaboo,iva0} or 2D trajectories~\cite{dragdiffusion,tora,draganything,simulatemotion,dragnuwa}. Later works introduced warping-based control~\cite{gowiththeflow,objctrl} and sparse point trajectory prompting~\cite{motionprompting,magicmotion}, enabling more expressive motion specification. In addition, 3D motion trajectories~\cite{objctrl,motionctrl,3dtrajmaster,cameractrl} have been explored to provide depth-aware controllability, and recent work further extends this direction by supporting heterogeneous motion inputs~\cite{anyi2v}. Motion transfer~\cite{ditflow,motionclone,sma,DeT,ropecraft} offers another practical route for controllable generation, but relies on reference videos and therefore does not allow users to directly specify motion.
\subsubsection{Training-free Video Motion Control.}
While motion control can be achieved by fine-tuning pretrained video models, the high retraining cost has driven growing interest in training-free approaches. Methods supporting user-defined motion typically induce motion by modifying the input latent, either through latent warping~\cite{freetraj,peekaboo} or by optimizing the latent with bounding-box similarity losses on intermediate features~\cite{smm,sgi2v,moft}. Another line of work focuses on motion transfer, matching motion representations from attention blocks to those of a reference video~\cite{motionclone,motionshop,ditflow}. Recent motion-transfer approaches further manipulate attention components by injecting reference queries~\cite{motionbyquery,uniedit} or modifying rotary position embeddings (RoPE)~\cite{ropecraft}. Unlike these methods, which inherit attention correspondences from a reference video, we explicitly warp queries to construct such correspondences for the desired motion. To the best of our knowledge, few prior works directly manipulate DiT attention to generate videos from a single image while following user-defined object or camera motion.
\begin{figure}[t!]
    \centering
    \includegraphics[width=\columnwidth]{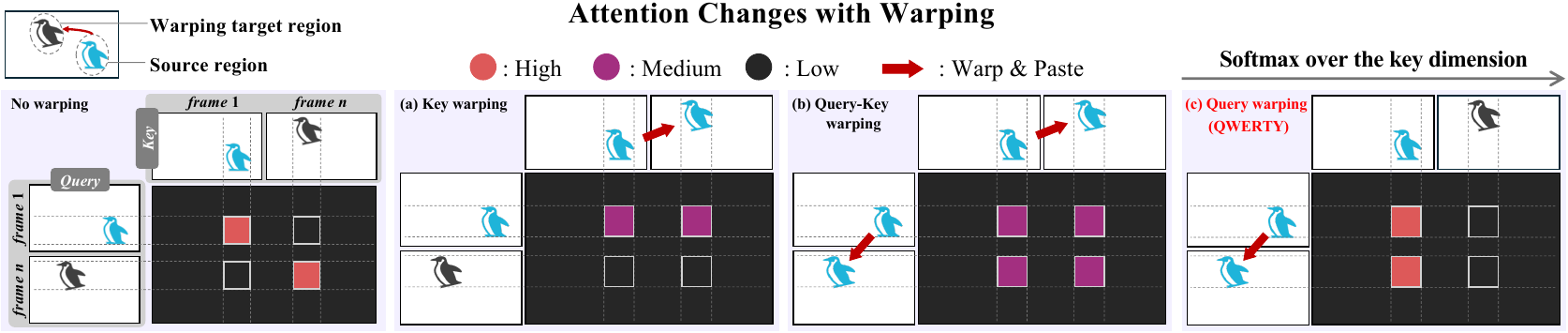}
    \vspace{-12pt}
    \caption{
    A conceptual illustration of how query warping guides motion. Warping takes tokens from the object's source region in frame 1 and pastes them at warped positions in frame $n$. Because the softmax is applied over keys, warping only the keys (a) or both queries and keys (b) does not establish a clear correspondence between the source and target regions. In (c), query warping concentrates attention from the target region in frame $n$ onto the source region in frame 1, which carries reliable semantic features.
    % A conceptual illustration of how query warping guides motion. Warping takes tokens from the object's source region in frame 1 and places them at warped positions in frame $n$. In (c), query warping concentrates attention from the target region in frame $n$ onto the source region in frame 1, which carries reliable semantic features. Because the softmax is applied over keys, warping only the keys (a) or both queries and keys (b) does not establish a clear correspondence between the source and target regions.
    % 우리가 query만을 와핑하는 이론적 직관에 대한 detailed illustration. warping이란 frame 1의 object source region의 토큰을 원하는 대로 warping하여 frame n에 붙여넣는 것을 의미한다. (b) query를 warping한 경우 frame n의 target region의 attention이 첫 프레임 (refence임) source region으로 집중되어 모션이 가이드된다. softmax가 key 방향으로 연산되기 때문에 key를 와핑한 경우 (c)와 query, key를 모두 와핑한 경우 (d)는 source, target region사이의 명확한 연관성을 만들지 못한다.
    % Limitations of displacement-based motion transfer.
    % Limitations of displacement-based motion transfer.
    }
    \label{fig:fig4}
    % \vspace{-9pt}
\end{figure}

\section{Why Query Warping?}
\label{sec:why_query}
This section analyzes why warping only the queries, among the attention components (queries, keys, and hidden states), is the most effective choice. Based on the observation in Fig.~\ref{fig:fig_2}(a), we consider the first frame to be the most stable feature map during early denoising steps, and therefore a reliable reference. Thus, the goal is to increase the attention between the target region in frame $n$ and the source region in frame 1. This encourages the target region to receive value vectors from the first-frame source region, thereby guiding the target motion.
Fig.~\ref{fig:fig4} illustrates how attention changes under three different warping strategies:

\myparagraph{(a) Key warping.} When keys are warped, the target-region query in frame~$n$ fails to attend strongly to any key, resulting in a dispersed attention pattern. Moreover, the attention of the source-region query in frame~1 is also dispersed, yielding no beneficial effect. Since softmax is applied over the key dimension independently for each query, modifying keys is not suitable for concentrating the attention of multiple queries onto a single meaningful source region.
% \myparagraph{Key warping.} When keys are warped, attention from a query becomes dispersed over multiple keys. In particular, the target-region query in frame~$n$ attends only weakly to both the source-region keys in frame~1 and the target-region keys in frame $n$, resulting in little useful guidance. Since the softmax is applied over the key dimension independently for each query, modifying keys alone cannot easily concentrate the attention of many queries on a single meaningful source region.
% \myparagraph{Key warping.} For a given query, attention becomes dispersed over multiple keys. In particular, the target-region query in frame $n$ has weak attention to both the source-region keys in frame 1 and the target-region keys in frame $n$, providing little useful guidance. Since the softmax is applied over the key dimension independently for each query, modifying keys alone cannot easily concentrate many queries on a single meaningful source region.

\myparagraph{(b) Query–Key warping (hidden-state warping).} Since queries and keys are derived from the same hidden state, jointly warping them behaves similarly to warping the hidden state itself. Although the attention between the target-region query in frame $n$ and the source-region key in frame 1 increases, attention is still spread out, similar to the key-warping case, and guidance remains diluted.
% \myparagraph{Query/key (hidden-state) warping.} Since queries and keys are both derived from the same hidden state, warping both is equivalent to warping the hidden state itself. Although the attention between the target-region query in frame $n$ and the source-region key in frame 1 increases, attention is still spread out, similar to the key-warping case, and guidance remains diluted.

% \myparagraph{Query warping (\MODELNAME).} Warping only the target-region queries in frame $n$ toward the source region in frame 1 yields strong attention between those queries and the first-frame source-region keys. This most effectively pulls the target-region features toward the reference features and provides clear motion guidance.
\myparagraph{(c) Query warping (\MODELNAME).} Warping the source-region queries from the first frame to the target region in frame~$n$ yields strong attention between those queries and the first-frame source-region keys. Thus, warping only the queries pulls the target-region features toward the reference features, providing clear motion guidance. Our results in Sec.~\ref{sec:ablation} empirically support this design choice.

This intuition can be formalized via the attention logits under a mild self-correspondence assumption. Motion control requires a target token \(i\) in frame \(n\) to attend to a source token \(s\) in frame 1. The target-to-source attention is \(\alpha_{i,s} = \exp(q_i^\top k_s/\sqrt{d}) / \sum_j \exp(q_i^\top k_j/\sqrt{d})\). Assuming that the source query aligns most with its own key, \ie, the self-correspondence score \(c_{\mathrm{self}} \triangleq q_s^\top k_s\) satisfies \(c_{\mathrm{self}} > q_s^\top k_j,\ \forall j \neq s\), query warping sets \(q_i \leftarrow q_s\), so \(q_i^\top k_s \approx c_{\mathrm{self}}\), directly increasing the numerator. In contrast, key warping only replaces target keys with source-like keys, \eg, \(\tilde{k}_t \leftarrow k_s\), leaving the numerator \(q_i^\top k_s\) unchanged; thus, \(\alpha_{i,s}\) need not increase. Query-key warping sets \(q_i \leftarrow q_s\), like query-only warping, but also replaces target keys with source-like keys, \(\tilde{k}_t \leftarrow k_s\). Thus, query-key warping also increases the numerator, but adds strong competing denominator terms with scores near \(c_{\mathrm{self}}\), diluting attention to the source key \(k_s\), unlike query-only warping. This formalizes why query warping most directly increases target-to-source attention and induces the desired correspondence.

\section{Methods}
\begin{figure*}[t!]
    \begin{center}
        \includegraphics[width=\textwidth]{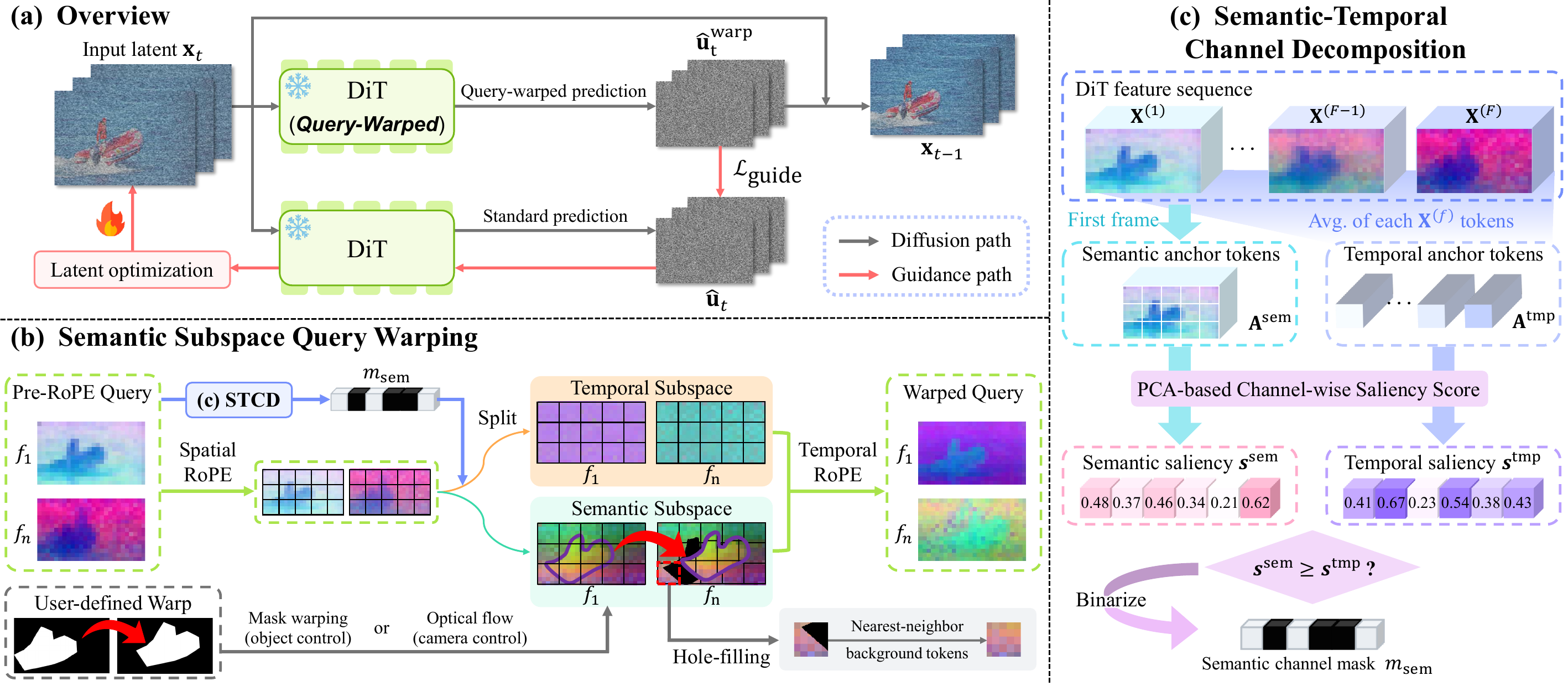}
    \end{center}
    \vspace{-12pt}
    \caption{
    Pipeline overview.
    {(a)} We perform diffusion steps using motion-inducing noise predicted from the query-warped video DiT. This query-warped noise is also used as a guidance signal to optimize the input latent, further stabilizing control.
    {(b)} We isolate frame-consistent semantic channels from the queries and warp them together with only the spatial axis of the 3D RoPE, delicately reshaping the attention distribution to induce motion without excessive distortion.
    {(c)} To separate semantic and temporal components in the query features, we define semantic anchor tokens and temporal anchor tokens from the query tokens, perform PCA on each anchor set to obtain channel-wise saliency scores, and then split the channels into two subspaces based on these scores.
    % (a) 우리는 video DiT의 query warping를 warping하여 얻은 motion-유발된 noise로 diffusion step을 진행한다. 이 query-warped noise는 input latent를 최적화하는 가이드로도 사용되어 control 안정성을 더욱 높인다. (b) 우리는 쿼리에서 frame-consistent한 semantic 채널만 분리하고 3D RoPE의 spatial 축만 함께 warping하여 과도한 왜곡 없이 어텐션 분포를 모션이 유발되도록 섬세하게 조작한다. (c) query 피쳐의 semantic-temporal 성분 분리를 위해 우리는 query 토큰들을 기반으로 semantic anchor token과 temporal anchor 토큰을 정의한다. 각 앵커 토큰을 가지고 PCA를 수행하여 채널 별 saliency score를 구하고, 이를 기준으로 채널을 두 subspace로 나눈다.
    % Pipeline overview \ym{Warping 부분 바꿔야되는데}
    }
    % \vspace{-8pt}
    \label{fig:fig_3}
\end{figure*}

In this section, we present \MODELNAME, as outlined in \cref{fig:fig_3}.
First, we introduce STCD (Sec. \ref{sec:channel_selection}). Second, we detail the query-warping procedure, including the application of STCD to queries (Sec. \ref{sec:query_warping}). Finally, we describe the motion-guided generation pipeline that performs diffusion with the query-warped DiT and latent optimization (Sec. \ref{sec:motion_guidance}). Although we describe our method using Wan 2.2 \cite{wan}, it can be applied to standard video DiTs with similar 3D full attention.
% 이 섹션에서는, we detail our \MODELNAME. first, 입력 피쳐 채널을 semantic subspace와 temporal subspace로 분리하는 STCD를 소개한다. 두번째로 query에 STCD를 적용하고 warp하는 자세한 수행과정을 설명한다. 마지막으로, query-warped DiT로 diffusion step을 수행하고 latent를 optimize하는 전반적인 motion-guided 비디오 생성 과정을 설명한다. 모든 과정은 Wan 2.2 I2V 모델을 기준으로 서술되어 있으며 대부분의 Video DiT에 범용적으로 적용 가능하다.

\subsection{Semantic-Temporal Channel Decomposition}
\label{sec:channel_selection}
% \subsection{Frame-invarient Semantic Subspace Filtering}
STCD is a PCA-driven disentanglement technique that isolates a frame-consistent semantic subspace from DiT features whose channels are entangled with temporal dynamics (Fig.~\ref{fig:fig_2}(a), second row). We first define two anchor token sets: \textit{semantic anchors}, whose within-frame spatial structure emphasizes semantic content, and \textit{temporal anchors}, whose across-frame variability highlights temporal change. Running PCA on each anchor set yields channel-wise saliency scores. We then compare the scores for each channel and apply query warping only where semantic saliency exceeds temporal saliency (Sec.~\ref{sec:query_warping}). Operationally, STCD defines a function that, given a feature tensor, returns a mask selecting semantically salient channels. The overall procedure is illustrated in Fig.~\ref{fig:fig_3}(c).
% \cy{여기는 3.1절에 대한 intro인거같은데 3.2절 언급이 나오면 안될듯. 3.1에서 설명할 것만 간략하게 소개하고 구조화.}
% \cy{STCD는 00을 하기 위한 또는 00을 가능하게해 뭐를 해결하기 위한 method이다. STCD는 PCA 뭐시기에 기반해 동작하며 우리는 STCD를 00어케 사용하기 위해 00토큰들을 활용한다. STCD는 다음 단계를 따른다. (i) 00 (ii) AA c BB. Let us first describe PCA-based Channel Saliency.}
% STCD는 프레임 variant 성분이 혼합된 DiT feature (Fig. \ref{fig:fig_2}(a) 상단)로부터 프레임 across로 균일한 semantic 성분만을 추출하기 위한 PCA 기반 테크닉이다. 이를 위해 우선, semantic 성분이 두드러져야 잘 구별될 수 있는 토큰들인 \textit{semantic anchor} tokens과 temporal (frame-variant) 성분이 두드러져야 잘 구별될 수 있는 토큰들인 \textit{temporal anchor} tokens을 정의한다. 이 두 토큰 집합을 가지고 각각 PCA를 수행하여 각 성분에 대한 channel-wise saliency score를 계산한다. 채널별로 두 스코어를 비교하여 semantic saliency score가 더 큰 채널에만 Sec. \ref{sec:query_warping}에서 query warping을 적용한다. 따라서 기술적?으로 STCD의 목적은 피쳐를 입력받았을 때 semantic salient한 채널의 mask를 출력하는 함수를 정의하는 것이며, 전반적인 과정은 Fig. \ref{fig:fig_3}(c)에 묘사되어 있다.

% \myparagraph{Notations.} Let $\mathbf{X} \in \mathbb{R}^{T \times D}$ be the per-token feature matrix, where $D$ is the channel dimension and $T = F \cdot H \cdot W$ is the total number of tokens obtained from $F$ video frames with $H \cdot W$ spatial tokens per frame.
% We reshape $\mathbf{X}$ into frame-wise token matrices
% $\mathbf{X}^{(f)} \in \mathbb{R}^{P \times D}$ for $f = 1,\dots,F$.

% \myparagraph{Channel Saliency Score.} 
% \myparagraph{PCA-based Channel Saliency.} 
\subsubsection{PCA-based Channel Saliency.} 
We begin by explaining how, given an arbitrary anchor tokens $\mathbf{A}\in\mathbb{R}^{N\times D}$ with $N$ tokens and $D$ channels, we compute a channel-wise \emph{discriminativeness} score.
Intuitively, a channel is discriminative for $\mathbf{A}$ if its values vary systematically across tokens (\eg, across spatial locations within a frame or across frames). 
PCA summarizes this cross-token variation: components with higher variance are more informative for distinguishing tokens.

Concretely, we center $\mathbf{A}$ across tokens and take its singular value decomposition (SVD), $\overline{\mathbf{A}}=\mathbf{U}\mathbf{\Sigma}\mathbf{V}^\top$, where the columns of $\mathbf{V}$ are channel loadings and the diagonal entries of $\mathbf{\Sigma}$ are the corresponding singular values $\sigma_i$. 
% We then assign each channel $d$ a saliency by weighting its loading magnitude by the corresponding singular values over the top $k$ components, taking the maximum so that a channel is considered salient if it strongly contributes to any dominant variation:
We then assign each channel $d$ a saliency by weighting its loading magnitude by the corresponding singular values over the top $k$ components:
% We then assign each channel $d$ a saliency by combining loading magnitude with variance explained over the top $k$ components:
{\small
\begin{equation}
s_d(\mathbf{A},k)
=
\max_{1\le i\le k}\,\sigma_i\,\bigl|{V}_{d,i}\bigr|,
\qquad d=1,\ldots,D.
\label{eq:channel-saliency}
\end{equation}
}%
This yields a score vector $\boldsymbol{s}(\mathbf{A},k)\in\mathbb{R}^D$ that identifies, for the given anchor token set, the channels most responsible for dominant token-wise variation. We compute $\boldsymbol{s}$ for both the semantic and temporal anchor sets and compare them per channel to label channels as semantic or frame-variant.

% ##############################################################################
% \myparagraph{Semantic Channel Saliency Score.}
% \myparagraph{Semantic Channel Saliency.}
\subsubsection{Semantic Channel Saliency.}
% We treat the first frame’s tokens as the \emph{semantic anchor} set, $\mathbf{A}^{\text{sem}}=\mathbf{X}^{(1)}\in\mathbb{R}^{P\times D}$.
We treat the first frame’s token features, $\mathbf{X}^{(1)}\in\mathbb{R}^{P\times D}$, where $P$ denotes the number of tokens in a single frame, as the \emph{semantic anchor} set, \ie, $\mathbf{A}^{\text{sem}}=\mathbf{X}^{(1)}$.
In DiT-based I2V models, first-frame features already exhibit a clear, well-structured layout even at early denoising steps (Fig.~\ref{fig:fig_2}(a)). This variation across spatial locations provides strong semantic cues, making $\mathbf{X}^{(1)}$ a suitable semantic anchor. We apply the saliency function in Eq.~\eqref{eq:channel-saliency} to obtain the semantic channel score vector
{\small
\begin{equation}
\boldsymbol{s}^{\text{sem}}
=
\boldsymbol{s} \bigl(\mathbf{A}^{\text{sem}},\,k_{\text{sem}}\bigr)
\in \mathbb{R}^D.
\label{eq:sem-score}
\end{equation}
}%
Here, $k_{\text{sem}}$ denotes the number of principal components.
% \cy{$\boldsymbol{s}$가 silancy func?$\boldsymbol{s}^{\text{sem}}$도 정의가 안된듯.. score vector겟지만}
% Here, $k_{\text{sem}}$ denotes the number of principal components used for scoring.
% Here, $k_{\text{sem}}$ denotes the number of principal components used for scoring (we use $k_{\text{sem}}=3$).

% \myparagraph{Temporal Channel Saliency.}
\subsubsection{Temporal Channel Saliency.}
To capture channels that vary across frames, we form \emph{temporal anchor} tokens by averaging over spatial positions within each frame. This suppresses within-frame spatial idiosyncrasies while preserving across-frame trends. Concretely, we compute
$\mathbf{a}^{\text{tmp}}_f=\tfrac{1}{P}\sum_{p=1}^{P}\mathbf{X}^{(f)}_{p,:}\in\mathbb{R}^{D}$
for each frame $f$, and stack them into
$\mathbf{A}^{\text{tmp}}\in\mathbb{R}^{F\times D}$ with rows $\mathbf{A}^{\text{tmp}}_{f,:}=\mathbf{a}^{\text{tmp}}_f$.
Applying the same scoring operator yields the temporal channel saliency vector
{\small
\begin{equation}
\boldsymbol{s}^{\text{tmp}}
=
\boldsymbol{s} \bigl(\mathbf{A}^{\text{tmp}},\,k_{\text{tmp}}\bigr)
\in\mathbb{R}^{D}.
\label{eq:tmp-score}
\end{equation}
}
% We use $k_{\text{tmp}}=3$ in all experiments.

% \myparagraph{Frame-consistent Semantic Channel Masking.}
\subsubsection{Frame-consistent Semantic Channel Masking.}
We identify channels that are more semantically salient than temporally salient by comparing the two saliency scores channel-wise (Eqs.~\eqref{eq:sem-score} and \eqref{eq:tmp-score}). Specifically, we form a semantic channel mask
$m_{\text{sem},d}=\mathbbold{1}[s^{\text{sem}}_{d}\ge s^{\text{tmp}}_{d}].$
To restrict query warping to semantic channels, we define a function that maps a feature sequence $\mathbf{X}$ to the binary semantic channel mask:
{\small
\begin{equation}
\boldsymbol{m}_{\text{sem}}
\;=\;
\mathrm{STCD} \bigl(\mathbf{X};\,k_{\text{sem}},k_{\text{tmp}}\bigr)
\in \{0,1\}^{D}.
\label{eq:stcd}
\end{equation}
}

% We identify channels that are more semantically descriptive than temporally variant
% by comparing the two saliency scores channel-wise.
% 우리는 semantic saliency score (\eqref{eq:sem-score}와 temporal saliency score (\eqref{eq:tmp-score})를 channel-wise로 비교해서 semantic channel mask $m_{\text{sem},d}=\mathbb{1}[\,s^{\text{sem}}_{d}\!\ge s^{\text{tmp}}_{d}\,]$를 얻는다. 우리는 restrict query warping to semantic channels 하기 위해, DiT feature를 입력받아 semantic channel mask를 출력하는 함수를 정의한다:
% {\small
% \begin{equation}
% \boldsymbol{m}_{\text{sem}}
% =
% \mathsf{STCDMask}\!\bigl(\mathbf{X};\,k_{\text{sem}},k_{\text{tmp}}\bigr)
% \in \{0,1\}^{D},
% \label{eq:stcd-mask}
% \end{equation}
% }

% ##############################################################################
\subsection{Semantic-Subspace Query Warping}
\label{sec:query_warping}
% Here, we describe the query warping procedure together with STCD 그리고 warping 도중에 발생하는 구멍을 다루는 전략에 대해서. 그 전에, 원활한 이해를 위해 3D RoPE에 대한 간단한 리뷰를 한다.
% Here, we describe the query warping procedure together with STCD, as well as the strategy for handling holes that arise during warping. Before presenting these details, we briefly review 3D RoPE for clarity.
We first briefly review 3D RoPE for clarity, and then describe the query warping procedure, along with our strategy for handling holes that arise during warping.
% Here, we detail the query warping procedure together with STCD and explain why warping queries within attention is uniquely capable of guiding motion.
% \cy{앞 section과 연결 안됨. 필요성, motivation 간략히 추가}
% This section details how we warp queries and how that process guides motion.
% 이 섹션에서는 query를 warping의 과정과 이를 통해 motion을 가이드하는 메커니즘을 구체화한다.
% 이 섹션에서는 query를 warping의 과정과 이를 통해 motion을 가이드하는 메커니즘을 구체화하며, 왜 hidden state를 warping하는 것보다 이점이 있는지에 대한 직관을 설명한다.

% \myparagraph{Preliminary: 3D-RoPE of video DiTs.}
\subsubsection{Preliminary: 3D RoPE of video DiTs.}
Rotary positional embedding (RoPE) \cite{rope} injects relative positional information into self-attention by rotating query and key feature dimensions. After rotation, the attention dot product depends on relative displacement.
For video, 3D RoPE treats each token as \((x,y,t)\). It splits the channel dimension into \([D_x \,\|\, D_y \,\|\, D_t]\), applies a separate 1D RoPE to each block, and then concatenates the rotated blocks.
Because these 1D RoPEs are applied to disjoint channel blocks, we can apply the spatial and temporal RoPE separately in our query warping.

\subsubsection{Query Warping Pipeline.}
As illustrated in Fig.~\ref{fig:fig_3}(b), our query-warping pipeline proceeds in three steps. 
\textbf{(\romannumeral 1)} From the pre-RoPE queries, we compute the frame-consistent semantic channel mask via STCD (Eq.~\eqref{eq:stcd}). 
\textbf{(\romannumeral 2)} We apply the spatial RoPE to the queries and warp the first-frame queries onto later frames according to the desired motion, pasting only the channels selected by the STCD mask. By performing warping after spatial RoPE, the queries in frame $n$ within the warped mask region share the same spatial positional encoding as the corresponding first-frame keys, thereby strengthening their spatial affinity in attention. Applying temporal RoPE jointly during warping would further increase their attention scores; however, this would encode the mask region in frame $n$ with the temporal position of frame 1, leading to excessive distortion of the attention distribution. 
\textbf{(\romannumeral 3)} Finally, we apply the temporal RoPE to the warped queries and proceed with the standard attention and denoising steps. 
Through this carefully designed query-warping strategy, we guide motion without disrupting the internal dynamics of the pretrained model.
% As illustrated in Fig.~\ref{fig:fig_3}(b), our query warping pipeline proceeds in three steps: \textbf{(\romannumeral 1)} From the pre-RoPE queries, we compute the frame-consistent semantic channel mask via STCD (Eq.~\eqref{eq:stcd}). \textbf{(\romannumeral 2)} We apply the spatial RoPE to the queries and warp the first-frame queries onto later frames 원하는 모션에 맞게, pasting only the channels selected by the STCD mask. By warping after spatial RoPE, the queries in frame $n$ within the warped first-frame mask region are placed in the same rotated basis as the first-frame keys in that mask, which their 어텐션에서 그들 사이의 공간적 연관성을 강화한다. temporal RoPE까지 같이 warping하게 되면 둘 사이의 어텐션이 더욱 높아지겠지만, frame n의 mask 영역이 frame 1의 temporal 순번?으로 인코딩되기 때문에 어텐션 분포를 과도하게 왜곡하게 된다. \textbf{(\romannumeral 3)} 마지막으로, We then apply the temporal RoPE to the warped queries and continue with the standard attention and denoising steps. 이렇게 세심한 query warping을 통해 우리는 모델의 내부를 망치지 않으면서도 원하는 모션을 가이드할 수 있다.

% 추가적으로, we 아래에 detail the hole-handling strategies during warping, which further improves control fidelity and suppresses artifacts. 이에 대한 더 구체적인 분석은 Suppl에 있다.

\subsubsection{Handling Warping Holes.}
We detail the hole-handling strategies during warping, which improve control fidelity and suppress visual artifacts. Additional analysis is provided in the Supplementary Material.

% \myparagraph{Warping with a mask (object control).}
\noindent \textbf{(i) Warping with a mask (object control).}
% \noindent \scalebox{0.8}{$\bullet$}~\textbf{  Warping with a mask (object control).}
Simply pasting the warped mask region leaves residual object traces at the source location; we therefore explicitly handle the source-region hole. On the first frame, we fill the hole with nearest-neighbor background tokens and paste both the warped mask region and the filled-hole region into each target frame (Fig.~\ref{fig:fig_3}(b)). This suppresses cloning artifacts that would otherwise leave a duplicate at the source location.
% To ensure that the object truly disappears from its original location, we do more than just paste the warped mask region. On the first frame, we also fill the mask hole with nearest-neighbor background tokens and paste both the warped mask region and the hole-fill into each target frame. This reduces cloning artifacts where the object would otherwise replicate at its source location.

% \myparagraph{Warping with optical flow (camera control).}
% \noindent \scalebox{0.8}{$\bullet$}~\textbf{  Warping with optical flow (camera control).}
\noindent \textbf{(ii) Warping with optical flow (camera control).}
Since optical flow is defined over the full spatial domain of each frame, warping is applied to the entire frame. Unlike the mask case, flow-based forward warping inherently creates holes on the target sampling grid due to incomplete coverage. Camera motion can also reveal regions not visible in the first frame, producing additional holes. Filling these holes with background tokens (akin to backward warping) propagates neighboring values into empty regions, causing stretching artifacts. Instead, we do not paste warped queries at hole locations; we leave the corresponding target-frame queries unchanged and allow the model to inpaint them naturally.

\subsection{Motion-guided Generation via Query Warping}
\label{sec:motion_guidance}
In this section, we describe the overall framework that integrates the components from Secs.~\ref{sec:channel_selection} and \ref{sec:query_warping} to control video motion through query-warped diffusion sampling and latent optimization, as illustrated in Fig.~\ref{fig:fig_3}(a).

\subsubsection{Query-Warped Prediction.}
We describe how query warping guides the I2V DiT under flow-matching formulation \cite{flow_matching} and note that the same principle applies to standard diffusion-based sampling. Let $\mathbf{c}$ denote the input image condition and $\mathbf{x}_t$ the latent at timestep $t$. The DiT, parameterized by $\theta$, predicts the velocity field. We denote the \emph{standard} velocity prediction by
\begin{equation}
\small
\hat{\mathbf{u}}_t = f_\theta(\mathbf{x}_t, t, \mathbf{c}).
\label{eq:standard_pred}
\end{equation}
The latent state is then advanced by an Euler step of the underlying flow:
\begin{equation}
\small
\mathbf{x}_{t-\Delta t} = \mathbf{x}_t + \Delta t \,\hat{\mathbf{u}}_t.
\label{eq:flow_update}
\end{equation}

To incorporate user-defined motion, we apply the query-warping procedure of Sec.~\ref{sec:query_warping} within the DiT attention blocks. Let $f_\theta^{\text{warp}}(\mathbf{x}_t, t, \mathbf{c}; \mathcal{W})$ denote a forward pass in which queries are warped according to the mask or flow field $\mathcal{W}$. This yields the \emph{query-warped} prediction:
\begin{equation}
\small
\hat{\mathbf{u}}_t^{\text{warp}}
    = f_\theta^{\text{warp}}(\mathbf{x}_t, t, \mathbf{c}; \mathcal{W}).
\label{eq:warp_pred}
\end{equation}
At early timesteps (i.e., the first five denoising steps), we use $\hat{\mathbf{u}}_t^{\text{warp}}$ in place of $\hat{\mathbf{u}}_t$ in Eq.~\eqref{eq:flow_update}, biasing the trajectory toward the desired motion.

\subsubsection{Self-guided Latent Optimization.}
While sampling with the query-warped prediction already provides strong motion guidance (Fig.~\ref{fig:fig_2}(b)), we further leverage it as a self-guidance signal for latent optimization. By directly updating the input latent, we can more explicitly steer the diffusion trajectory toward the desired motion.
Specifically, at each guided timestep, we compute both the standard prediction $\hat{\mathbf{u}}_t$ and the query-warped prediction $\hat{\mathbf{u}}_t^{\text{warp}}$. Keeping the DiT parameters frozen, we update the latent $\mathbf{x}_t$ to match the standard prediction to the warped one. Gradients are stopped on the warped branch and propagated only through the standard branch (Fig.~\ref{fig:fig_3}(a)).
% Our method also leverages the query-warped diffusion noise as a self-guidance signal that refines the latent toward a motion-consistent manifold. Inspired by RoPECraft~\cite{ropecraft}, which guides motion using noise predicted from a reference video, we instead construct two branches for the same sample: a standard prediction $\hat{\mathbf{u}}_t$ and a query-warped prediction $\hat{\mathbf{u}}_t^{\text{warp}}$. We then update the latent $\mathbf{x}_t$ so that the standard branch aligns with the motion encoded in the warped branch, while keeping the DiT itself frozen.

% To stabilize this guidance, we adopt the phase-constraint idea of RoPECraft~\cite{ropecraft}, encouraging the two predictions to match in magnitude and in Fourier-domain phase, which encodes motion structure. Our self-guided loss is
To stabilize the guidance, we adopt the phase-constraint idea of RoPECraft \cite{ropecraft}, encouraging the standard and query-warped predictions to match in magnitude while aligning their Fourier phases, which capture motion structure.
We formulate the self-guided loss as
{\small
\begin{equation}
\begin{aligned}
\mathcal{L}_{\text{guide}}
&=
\|\hat{\mathbf{u}}_t - \hat{\mathbf{u}}_t^{\text{warp}}\|_2^2
+ \lambda\,\|\cos(\angle \mathcal{F}(\hat{\mathbf{u}}_t))
            - \cos(\angle \mathcal{F}(\hat{\mathbf{u}}_t^{\text{warp}}))\| \\
            % - \cos(\angle \mathcal{F}(\hat{\mathbf{u}}_t^{\text{warp}}))\|_2 \\
&\quad
+ \lambda\,\|\sin(\angle \mathcal{F}(\hat{\mathbf{u}}_t))
            - \sin(\angle \mathcal{F}(\hat{\mathbf{u}}_t^{\text{warp}}))\|
            % - \sin(\angle \mathcal{F}(\hat{\mathbf{u}}_t^{\text{warp}}))\|_2
\end{aligned}
\label{eq:self_guided_loss}
\end{equation}
}%
where $\mathcal{F}(\cdot)$ denotes the Fourier transform, $\angle$ extracts the phase angle, and $\lambda$ controls the weight of the phase constraint. Following RoPECraft, we set $\lambda=0.1$.

% Our self-guided loss is
% \begin{equation}
% \small
% \begin{aligned}
% \mathcal{L}_{\text{guide}}
% &=
% \|\hat{\mathbf{u}}_t - \hat{\mathbf{u}}_t^{\text{warp}}\|_2^2
% + \lambda\,\|\cos \mathcal{F}(\hat{\mathbf{u}}_t)
%             - \cos \mathcal{F}(\hat{\mathbf{u}}_t^{\text{warp}})\|_2 \\
% &\quad\;\;
% + \lambda\,\|\sin \mathcal{F}(\hat{\mathbf{u}}_t)
%             - \sin \mathcal{F}(\hat{\mathbf{u}}_t^{\text{warp}})\|_2
% \end{aligned}
% \label{eq:self_guided_loss}
% \end{equation}
% where $\mathcal{F}(\cdot)$ is the Fourier transform and $\lambda$ controls the phase weight. We set $\lambda = 0.1$ in all experiments, following RoPECraft. This self-guided refinement improves motion control and visual quality이지만, 상황에 따라 어느정도 생략 가능하다.

\begin{figure*}[t!]
    \begin{center}
        \includegraphics[width=\textwidth]{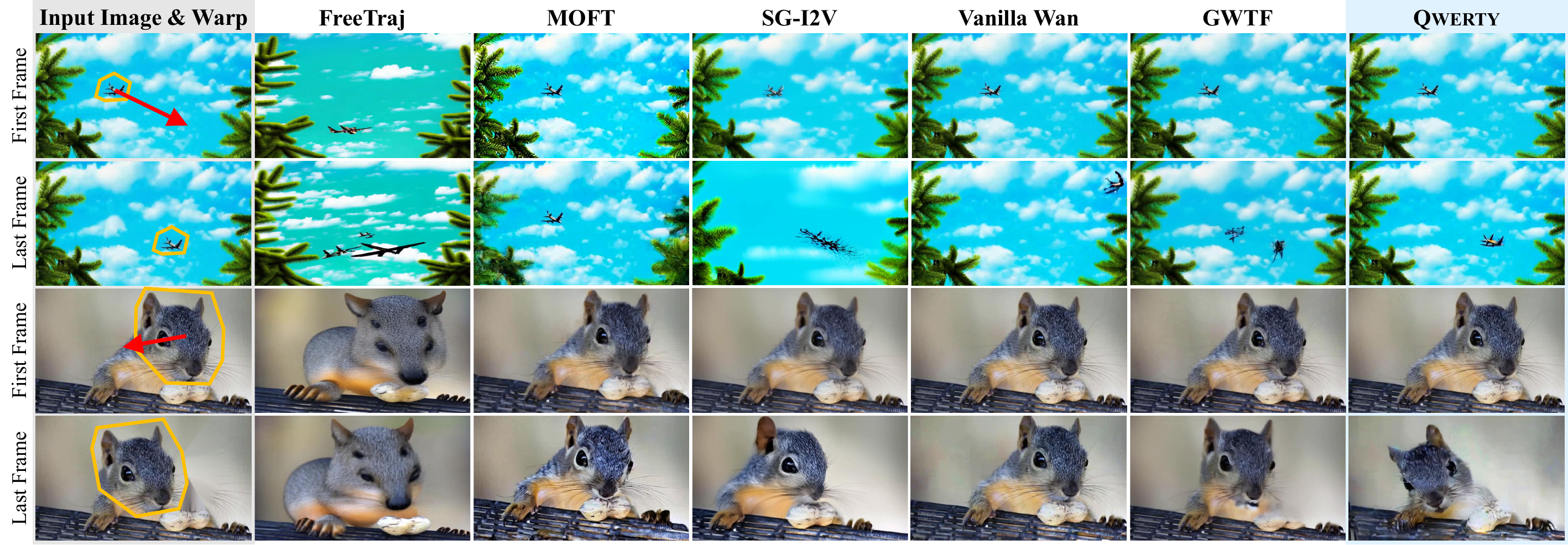}
    \end{center}
    \vspace{-12pt}
    \caption{Qualitative comparisons of local \textit{object} motion control. Polygonal object region masks and user-defined warps are given as input. 
    }
    \label{fig:fig_5}
    % \vspace{-8pt}
\end{figure*}
\begin{figure*}[t!]
    \begin{center}
        \includegraphics[width=\textwidth]{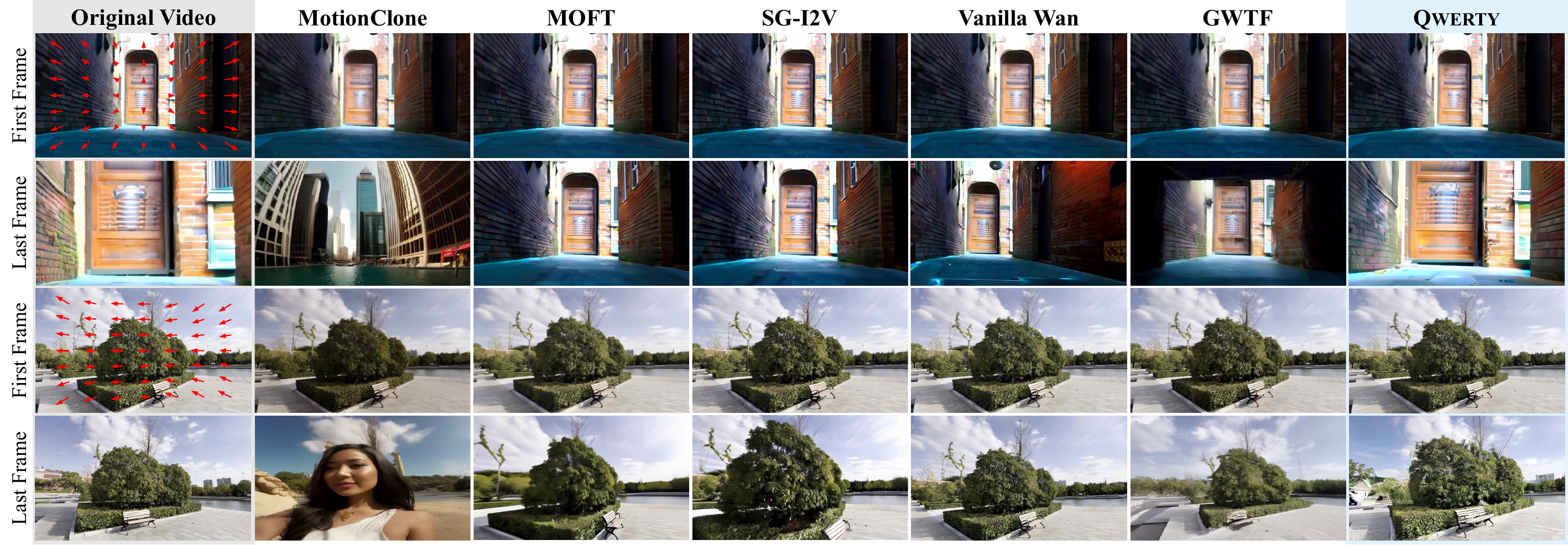}
    \end{center}
    \vspace{-12pt}
    \caption{Qualitative comparisons of \textit{camera} motion control. For evaluation, we use optical flow estimated from the original video. In practice, users can instead provide optical flow derived from a depth map and camera trajectory (Sec.~\ref{sec:ablation}).}
    \label{fig:fig_6}
    % \vspace{-5pt}
\end{figure*}

\section{Experiments}
\label{sec:experiments}
In this section, we present experiments to validate \MODELNAME. Additional results, analyses, and discussions are provided in the supplementary material.
% In this section, we provide extensive experiments to validate \MODELNAME. 추가적인 구현?디테일, 실험, 분석, 디스커션 등은 서플에 있다..

% \noindent \textbf{Evaluation metrics. }

% \input{sec/tables/main_table}

\subsection{Setup}

% \myparagraph{Implementation details.}
% \chu{}

\myparagraph{Baselines.}
We compare \MODELNAME with both U-Net– and DiT-based motion-control methods.
SG-I2V~\cite{sgi2v} and MOFT~\cite{moft} are training-free U-Net approaches that use bounding-box trajectories to control object or camera motion.
FreeTraj~\cite{freetraj}, originally designed for text-to-video (T2V), is adapted to the I2V setting and evaluated only for object control since it does not support camera motion.
MotionClone~\cite{motionclone}, a T2V motion-transfer method, is also adapted to I2V, where we provide the ground-truth video as motion reference to establish a strong U-Net baseline for camera control comparison.
% Since direct training-free counterparts for DiT-based I2V models remain limited, we adapt methods originally developed for U-Net models, including Noise Warping~\cite{freetraj} and SG-I2V, to the DiT backbone.
Since direct training-free counterparts for DiT-based I2V models are not readily available, we adapt methods developed for U-Net models, including Noise Warping~\cite{freetraj} and SG-I2V, to the DiT backbone.
We include Go-With-The-Flow (GWTF)~\cite{gowiththeflow}, a finetuned DiT model enabling motion control with the same mask and optical-flow conditioning interface as ours, serving as a strong upper-bound baseline.
Our main results are reported on Wan~2.2 TI2V-5B~\cite{wan}, with additional ablations on CogVideoX-I2V-5B to demonstrate cross-backbone effectiveness.
Implementation details for all baselines and our method are provided in the supplementary material.

\begin{table*}[t]
    \centering
    \caption{Quantitative comparisons with baselines. $\Uparrow$/$\Downarrow$ indicates better is higher/lower. \textbf{Best} and \underline{second best} are marked. \colorbox{MyBlue!70!white}{Light blue rows} denote our model. $^*$ indicates using the ground-truth video as a motion reference for motion transfer.}
    % \caption{Quantitative comparisons with baselines. $\Uparrow$/$\Downarrow$ indicates a higher/lower value is better. \textbf{Best} and \underline{second} are highlighted. \colorbox{MyBlue!70!white}{Light blue background rows} indicate our final model. Dashed lines separate datasets from baselines. $^*$~indicates using the ground-truth video as a motion reference video for the motion transfer method.}
    % \vspace{-8pt}
    
    \definecolor{Gray}{gray}{0.8}

    \resizebox{\linewidth}{!}{
    \begin{tabular}{
        c!{\vrule}c!{\vrule}c!{\vrule}c!{\vrule}ccccccc  % DiT | Method | Train | Backbone | 8 metrics
        % c!{\vrule}c!{\vrule}c!{\vrule}c!{\vrule}cccccccc  % DiT | Method | Train | Backbone | 8 metrics
    }
    \toprule
    % \noalign{\hrule height \heavyrulewidth}
    % ================= HEADER =================
    \multicolumn{2}{c!{\vrule}}{\multirow{2}{*}{\centering Method}}
    & \multirow{2}{*}{\begin{tabular}{@{}c@{}}Training\\free\end{tabular}}
    & \multirow{2}{*}{\centering Backbone}
    & \multirow{2}{*}{\centering FID $\Downarrow$}
    & \raisebox{-0.25ex}{FVD} $\Downarrow$
    & \multirow{2}{*}{\centering FTD $\Downarrow$}
    % & \multirow{2}{*}{\centering OF err. $\Downarrow$}
    & \multicolumn{4}{c}{VBench $\Uparrow$} \\ \cline{8-11}
    \multicolumn{2}{c!{\vrule}}{} &  &  &  & \raisebox{0.45ex}{(×10$^{3}$)} &  & SC & BC & MS & TF \\
    \hline\hline

    % ========= Local object (VIPSeg) =========
    \rowcolor{gray!20}
    \multicolumn{11}{c}{\hspace{0.45\linewidth}\textit{\textbf{Object motion control (VIPSeg)}}} \\

    % DiT X
    \multirow{3}{*}{\rotatebox[origin=c]{90}{\textit{U-N\large{et}}}}
    % & FreeTraj & \checkmark & VideoCrafter & 126.8 & 0.924 & 0.528 & 0.832 & 0.886 & 0.901 & 0.979 & 0.968 \\
    & FreeTraj & \checkmark & VideoCrafter & 126.8 & 0.924 & 0.528 & 0.886 & 0.901 & 0.979 & 0.968 \\
    % & MOFT     & \checkmark & SVD          & 76.04 & 0.861 & \underline{0.527} & 0.872 & 0.955 & 0.944 & 0.981 & 0.964 \\
    & MOFT     & \checkmark & SVD          & 76.04 & 0.861 & \underline{0.527} & 0.955 & 0.944 & 0.981 & 0.964 \\
    & SG-I2V   & \checkmark & SVD          & 60.83 & 0.798 & 0.579 & 0.952 & 0.961 & 0.974 & 0.946 \\
    % & SG-I2V   & \checkmark & SVD          & 60.83 & 0.798 & 0.579 & 0.860 & 0.952 & 0.961 & 0.974 & 0.946 \\
    \hdashline
    % DiT O
    \multirow{5}{*}{\rotatebox[origin=c]{90}{\textit{DiT}}}
      & \raisebox{-0.25ex}{Vanilla Wan}     & --          & Wan         & 46.17 & 0.677 & 0.534 & \underline{0.977}  & 0.973 & 0.983 & 0.973 \\
      % & \raisebox{-0.25ex}{Vanilla Wan}     & --          & Wan         & 46.17 & 0.677 & 0.534 & \underline{0.823} & \underline{0.977}  & 0.973 & 0.983 & 0.973 \\
      & Noise Warping      & \checkmark & Wan & 270.38 & 3.392 & 0.569  & 0.725  & 0.753 & 0.955 & 0.955 \\
      % & Noise Warping      & \checkmark & Wan & 270.38 & 3.392 & 0.569 & 0.834 & 0.725  & 0.753 & 0.955 & 0.955 \\
      & SG-I2V      & \checkmark & Wan & 107.98 & 0.704 & 0.600 & 0.976  & 0.960 & 0.962 & 0.967 \\
      % & SG-I2V      & \checkmark & Wan & 107.98 & 0.704 & 0.600 & 0.824 & 0.976  & 0.960 & 0.962 & 0.967 \\
      & \raisebox{-0.25ex}{GWTF}    &   \myxmark     & --            & \textbf{40.19} & \textbf{0.645} & 0.528 & 0.958 & \underline{0.975} & \textbf{0.993} & \textbf{0.988} \\
      % & \raisebox{-0.25ex}{GWTF}    &   \myxmark     & --            & \textbf{40.19} & \textbf{0.645} & 0.528 & \textbf{0.777} & 0.958 & \underline{0.975} & \textbf{0.993} & \textbf{0.988} \\
      % & \raisebox{-0.25ex}{GWTF$^{\;\!*}$}    &   \myxmark     & --            & \textbf{40.19} & \textbf{0.645} & 0.528 & \textbf{0.777} & 0.958 & \underline{0.975} & \textbf{0.993} & \textbf{0.988} \\
      % & FreeTraj & \checkmark & Wan          & 0.000 & \textbf{0.000} & 0.000 & 0.000 & 0.000 & 0.000 & 0.000 & 0.000 \\
      & \cellcolor{MyBlue!70!white}\MODELNAME
      & \cellcolor{MyBlue!70!white}\checkmark
      & \cellcolor{MyBlue!70!white}Wan
      & \cellcolor{MyBlue!70!white} \underline{45.95}
      & \cellcolor{MyBlue!70!white} \underline{0.667}
      & \cellcolor{MyBlue!70!white}\textbf{0.512}
      % & \cellcolor{MyBlue!70!white} \underline{0.823}
      & \cellcolor{MyBlue!70!white} \textbf{0.978}
      & \cellcolor{MyBlue!70!white}\textbf{0.981}
      & \cellcolor{MyBlue!70!white}\underline{0.985}
      & \cellcolor{MyBlue!70!white}\underline{0.977} \\
    \hline

    % ======== Camera movement (DL3DV) ========
    \rowcolor{gray!20}
    \multicolumn{11}{c}{\hspace{0.45\linewidth}\textit{\textbf{Camera motion control (DL3DV)}}} \\

    \multirow{3}{*}{\rotatebox[origin=c]{90}{\textit{U-N\large{et}}}}
    % & MotionClone$^{\;\!*}$ & \checkmark & AnimateDiff & 96.63 & 1.624 & 0.683 & \underline{0.613} & 0.897  & 0.934 & 0.969 & 0.931 \\
    & MotionClone$^{\;\!*}$ & \checkmark & AnimateDiff & 96.63 & 1.624 & 0.683  & 0.897  & 0.934 & 0.969 & 0.931 \\
    & MOFT        & \checkmark & SVD & 51.84 & 1.327 & 0.424 & 0.955    & 0.967 & 0.958 & 0.961 \\
    % & MOFT        & \checkmark & SVD & 51.84 & 1.327 & 0.424 & 0.752 & 0.955    & 0.967 & 0.958 & 0.961 \\
    & SG-I2V      & \checkmark & SVD & 45.18 & \underline{1.265} & 0.578 & {0.966}  & 0.973 & {0.979} & 0.936 \\
    % & SG-I2V      & \checkmark & SVD & 45.18 & \underline{1.265} & 0.578 & 0.702 & {0.966}  & 0.973 & {0.979} & 0.936 \\
    \hdashline
    \multirow{5}{*}{\rotatebox[origin=c]{90}{\textit{DiT}}}
      & \raisebox{-0.25ex}{Vanilla Wan}     & --          & Wan         & \underline{32.21} & 1.362 & \underline{0.367} & \textbf{0.977}  & \underline{0.974} & 0.978 & \underline{0.967} \\
      % & \raisebox{-0.25ex}{Vanilla Wan}     & --          & Wan         & \underline{32.21} & 1.362 & \underline{0.367} & 0.816 & \textbf{0.977}  & \underline{0.974} & 0.978 & \underline{0.967} \\
      & Noise Warping      & \checkmark & Wan & 282.93 & 5.663 & 0.455 & 0.719  & 0.747 & 0.972 & 0.966 \\
      % & Noise Warping      & \checkmark & Wan & 282.93 & 5.663 & 0.455 & \textbf{0.607} & 0.719  & 0.747 & 0.972 & 0.966 \\
      & SG-I2V      & \checkmark & Wan & 42.46 & 1.524 & 0.438 & \underline{0.976}  & 0.970 & \underline{0.981} & 0.941 \\
      % & SG-I2V      & \checkmark & Wan & 42.46 & 1.524 & 0.438 & 0.767 & \underline{0.976}  & 0.970 & \underline{0.981} & 0.941 \\
      & \raisebox{-0.25ex}{GWTF}        & \myxmark          & --   & \textbf{30.32} & 1.576 & 0.513 & 0.962 & \textbf{0.978} & \textbf{0.988} & \textbf{0.976} \\
      % & \raisebox{-0.25ex}{GWTF}        & \myxmark          & --   & \textbf{30.32} & 1.576 & 0.513 & 0.703 & 0.962 & \textbf{0.978} & \textbf{0.988} & \textbf{0.976} \\
      % & \raisebox{-0.25ex}{GWTF$^{\;\!*}$}        & \myxmark          & --   & \textbf{30.32} & 1.576 & 0.513 & 0.703 & 0.962 & \textbf{0.978} & \textbf{0.988} & \textbf{0.976} \\
      % & DiTFlow & \checkmark & CogVideoX & 0.000 & 0.000 & 0.000 & 0.000 & -     & 0.000 & 0.000 & 0.000 \\
      & \cellcolor{MyBlue!70!white}\MODELNAME
      & \cellcolor{MyBlue!70!white}\checkmark
      & \cellcolor{MyBlue!70!white}Wan
      & \cellcolor{MyBlue!70!white}35.56
      & \cellcolor{MyBlue!70!white} \textbf{1.007}
      & \cellcolor{MyBlue!70!white} \textbf{0.237}
      % & \cellcolor{MyBlue!70!white} {0.700}
      & \cellcolor{MyBlue!70!white}0.946
      & \cellcolor{MyBlue!70!white}0.956
      & \cellcolor{MyBlue!70!white}0.941
      & \cellcolor{MyBlue!70!white}0.935 \\
    \hline
    \end{tabular}
    }
    % \begin{flushleft}\vspace{-0.6em}
    % \footnotesize{\;\;$^*$~indicates methods that require additional training. \quad $^\dagger$~indicates using the ground-truth video as a motion reference video for the motion transfer method.}\end{flushleft}
    \label{tab:main}

    % \vspace{-10pt}
    
\end{table*}

\myparagraph{Datasets.}
% \ym{VIPSeg, DL3DV, custom object dataset (veo3, sora), camera(wan based)}
To evaluate both video quality and motion controllability, we conduct experiments on two datasets targeting object and camera motion control.
Following prior works~\cite{sgi2v, gowiththeflow}, object motion control is evaluated on 100 VIPSeg~\cite{vipseg} validation videos with dominant object motion, where mask-based warps are annotated.
Camera motion control is evaluated on 100 randomly sampled videos from DL3DV~\cite{dl3dv}.
Preprocessing and labeling details are provided in the supplementary material, and the selected video IDs will be released with the code.
\myparagraph{Evaluation metrics.}
We evaluate both generation quality and motion controllability of the generated videos.
For generation quality, we report Fréchet Inception Distance (FID)~\cite{fid}, Fréchet Video Distance (FVD)~\cite{fvd}, and four VBench metrics~\cite{vbench, vbench2}: subject consistency (SC), background consistency (BC), motion smoothness (MS), and temporal flickering (TF).
For motion controllability, we adopt the recent Fréchet Trajectory Distance (FTD)~\cite{ropecraft}, which measures alignment with the target motion for both object and camera movement.
For fair comparison, videos are generated at their native resolution and resized to 480$\times$832 for evaluation, and the frame count is standardized by truncating each video to the smallest maximum length supported by the compared models.
\begin{figure*}[t!]
    \begin{center}
        \includegraphics[width=\textwidth]{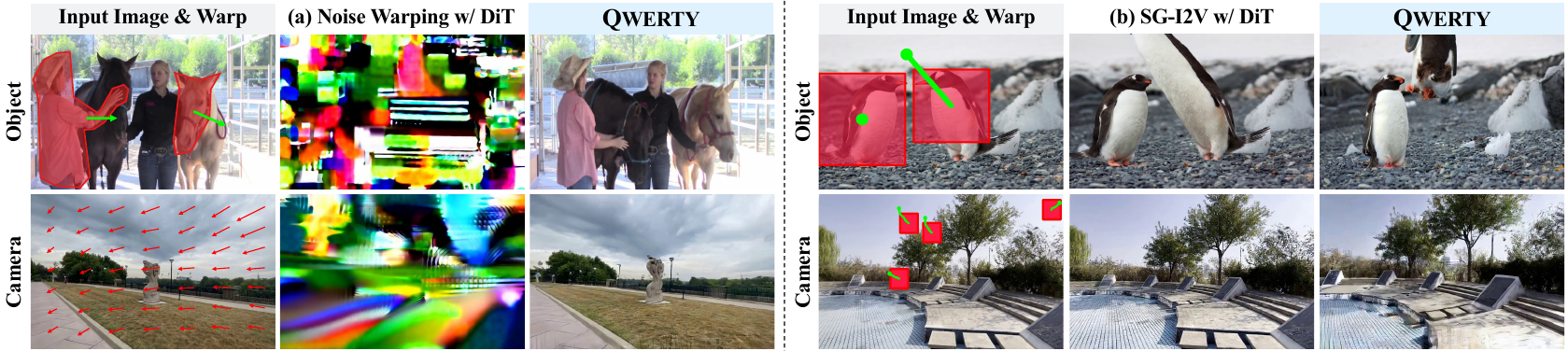}
    \end{center}
    \vspace{-12pt}
    \caption{Failure of U-Net–based motion-control methods on DiT. (a) Noise Warping collapses generation. (b) SG-I2V yields mostly static results, with occasional unintended motion. In contrast, \MODELNAME achieves reliable object and camera control on DiT.}
    \label{fig:fig_7}
    % \vspace{-9pt}
\end{figure*}

\subsection{Qualitative and Quantitative Results}
We present qualitative and quantitative comparisons with baselines in Fig.~\ref{fig:fig_5}, Fig.~\ref{fig:fig_6}, and Table~\ref{tab:main}. Fig.~\ref{fig:fig_7} shows training-free control methods designed for U-Net do not transfer well to DiT. Our results are mainly reported on Wan, while qualitative results on CogVideoX are provided in the supplementary material.
% Our main results are reported on Wan~2.2 TI2V-5B~\cite{wan}, with additional ablations on CogVideoX-I2V-5B to demonstrate cross-backbone effectiveness.

\myparagraph{Object control.}
% Fig. 5 illustrates the strong object-motion controllability of our method. Baseline methods show poor motion quality ~\cite{moft} and often duplicated or distorted target objects. Especially, generated videos lacked consistency with the input image ~\cite{freetraj}, turned the object controlling motion prompt into a camera movement ~\cite{sgi2v}. GWTF, which is a training method, seemed to follow the given input warp, but the generated motion was not 맥락에 맞아 and was highly distorted. Our method was the only one to generate high quality videos with successful motion compliance.
Fig.~\ref{fig:fig_5} illustrates strong object motion controllability of our method. 
Baseline approaches often exhibit limited motion quality, producing duplicated or distorted objects. Some methods fail to preserve consistency with the input image, while others incorrectly induce camera movement instead of object motion. 
Although GWTF, a training-based approach, tends to follow the control more strictly, the resulting motion is often contextually unnatural or distorted.
In contrast, our method generates high-fidelity videos with accurate motion compliance while producing semantically coherent motion. This behavior is more clearly observed in the supplementary videos.
Consistently, Table~\ref{tab:main} shows that our method achieves the best overall performance among training-free approaches and delivers video quality and control accuracy comparable to GWTF. Notably, despite manipulating internal components of the model for motion control, our approach improves visual fidelity compared to vanilla Wan.
% Fig.~\ref{fig:fig_5} illustrates the strong object motion controllability of our method.
% Baseline approaches exhibit limited motion quality, producing duplicated or distorted target objects. Some methods fail to preserve consistency with the input image, while others incorrectly generate a camera movement instead of object motion. Although GWTF, a training-based approach, tends to strictly follow the given control, the resulting motion is often contextually 부자연스럽거나 or distorted. In comparison, our method produces high-fidelity videos with accurate motion compliance (여기에 semantically 자연스러운 모션을 스스로 만든다고 해줘). 서플 비디오를 보면 이를 더 잘 느낄 수 있다.
% Reinforcing these observations, Table~\ref{tab:main} shows that our method achieves the best overall performance among training-free approaches and delivers video quality and control accuracy comparable to GWTF. Notably, despite manipulating the internal components of the model for motion control, our approach even improves visual fidelity compared to the original Wan model. U-Net-based models generally exhibited lower video quality than DiT-based models (VBench 값 참고).

\myparagraph{Camera control.}
\cref{fig:fig_6} shows that \MODELNAME aligns most faithfully with the instructed motion prompts, even under challenging conditions such as large zoom-ins or tilting camera motions. In contrast, other methods either fail to follow the specified motion or produce noticeable artifacts. In particular, GWTF performs poorly on DL3DV, a real-world scene dataset, suggesting that fine-tuning–based approaches may degrade the backbone’s pretrained generalization.
In Table~\ref{tab:main}, \MODELNAME outperforms nearly all baselines in camera-motion controllability, achieving substantial improvements in both FVD and FTD. Since generating camera motion inherently requires synthesizing novel scene content, VBench metrics such as SC or BC tend to trade off with camera-motion controllability. Methods such as vanilla Wan or SG-I2V adapted to DiT achieve higher VBench scores, as they often produce static or uncontrolled videos rather than true motion control. Notably, our method surpasses the training-based GWTF in both FVD and FTD, highlighting the effectiveness of our training-free design.
\begin{table}[t]
    \centering
    % \vspace{-8pt}

    % ---------- LEFT TABLE ----------
    \begin{minipage}[t]{0.480\linewidth}
        \centering
        \captionof{table}{Ablation results on Wan for camera control on the DL3DV dataset. $\Uparrow$/$\Downarrow$ indicates better is higher/lower. \textbf{Best} and \underline{second best} are highlighted.}
        % \captionof{table}{Ablation results showing the impact of our components on performance on the DL3DV dataset (Setting A). $\Downarrow$ indicates that lower is better. \textbf{Best} and \underline{second best} are highlighted.}
        \label{tab:ablation_A}
        \definecolor{Gray}{gray}{0.8}
        % \vspace{-8pt}

        \resizebox{\linewidth}{!}{
        \begin{tabular}{cllcccc}
            \toprule
            \multicolumn{3}{c}{\multirow{2}{*}{\centering Option}}
            & \raisebox{-0.25ex}{FVD} $\Downarrow$
            & \multirow{2}{*}{\centering FTD $\Downarrow$}
            & \multicolumn{2}{c}{VBench $\Uparrow$} \\ \cline{6-7}
            & & & \raisebox{0.45ex}{(×10$^{3}$)} & & BC & MS \\
            \hline\hline

            & & Vanilla Wan
                & 1.362 & 0.367 & \textbf{0.974} & \underline{0.978} \\
            \hline

            \multirow{6}{*}{\rotatebox[origin=c]{90}{\textit{Warping Target}}}
              & {(a)} & No STCD & 1.348 & 0.297 & 0.956 & 0.940 \\
              & {(b)} & \textit{w/o} spat. RoPE & 1.419 & 0.299 & 0.961 & \underline{0.978} \\
              & {(c)} & \textit{w/} temp. RoPE & 1.348 & 0.285 & 0.955 & 0.945 \\
              & {(d)} & $K$ & 1.482 & 0.293 & 0.969 & 0.861 \\
              & {(e)} & $Q, K$ & 1.427 & 0.290 & \underline{0.971} & \textbf{0.980} \\
              & {(f)} & Attn. output & 1.590 & 0.288 & 0.940 & 0.939 \\
            \hline

            \multirow{3}{*}{\rotatebox[origin=c]{90}{\textit{Optim.}}}
              & {(g)} & No opt. & \underline{1.301} & \underline{0.274} & 0.962 & 0.956 \\
              & {(h)} & Opt. only & 1.423 & 0.355 & 0.940 & 0.939 \\
              & {(i)} & \textit{w/o} phase loss & 1.440 & 0.285 & 0.949 & 0.942 \\
            \hline

            \rowcolor{MyBlue!70!white}
            & & \MODELNAME
                & \textbf{1.007} & \textbf{0.237} & {0.956} & {0.941} \\
            \noalign{\hrule height \heavyrulewidth}
        \end{tabular}
        }

        % \vspace{-8pt}
    \end{minipage}
    \hfill
    % ---------- RIGHT TABLE ----------
    \begin{minipage}[t]{0.505\linewidth}
        \centering
        \captionof{table}{Ablation results on CogVideoX for object control on the VIPSeg dataset. $\Uparrow$/$\Downarrow$ indicates better is higher/lower. \textbf{Best} and \underline{second best} are highlighted.}
        % \captionof{table}{Ablation results showing the impact of our components on performance on the DL3DV dataset (Setting B). $\Downarrow$ indicates that lower is better. \textbf{Best} and \underline{second best} are highlighted.}
        \label{tab:ablation_B}
        \definecolor{Gray}{gray}{0.8}
        % \vspace{-8pt}

        \resizebox{\linewidth}{!}{
        \begin{tabular}{cllcccc}
            \toprule
            \multicolumn{3}{c}{\multirow{2}{*}{\centering Option}}
            & \raisebox{-0.25ex}{FVD} $\Downarrow$
            & \multirow{2}{*}{\centering FTD $\Downarrow$}
            & \multicolumn{2}{c}{VBench $\Uparrow$} \\ \cline{6-7}
            & & & \raisebox{0.45ex}{(×10$^{3}$)} & & BC & MS \\
            \hline\hline

            & & Vanilla CogVideoX
                & \underline{0.669} & 0.537 & {0.974} & \underline{0.990} \\
            \hline

            \multirow{6}{*}{\rotatebox[origin=c]{90}{\textit{Warping Target}}}
              & {(a)} & No STCD & 0.697 & 0.542 & 0.961 & 0.987 \\
              & {(b)} & \textit{w/o} spat. RoPE & 1.578 & 0.531 & 0.835 & {0.973} \\
              & {(c)} & \textit{w/} temp. RoPE & 0.831 & 0.511 & 0.959 & 0.944 \\
              & {(d)} & $K$ & 0.759 & 0.544 & 0.966 & 0.928 \\
              & {(e)} & $Q, K$ & 0.727 & 0.531 & \underline{0.975} & {0.989} \\
              & {(f)} & Attn. output & 0.745 & 0.544 & 0.943 & 0.971 \\
            \hline

            \multirow{3}{*}{\rotatebox[origin=c]{90}{\textit{Optim.}}}
              & {(g)} & No opt. & \underline{0.669} & \underline{0.501} & 0.972 & 0.980 \\
              & {(h)} & Opt. only & 0.688 & 0.536 & 0.977 & 0.979 \\
              & {(i)} & \textit{w/o} phase loss & 0.704 & 0.514 & 0.964 & 0.955 \\
            \hline

            \rowcolor{MyBlue!70!white}
            & & \MODELNAME
                & \textbf{0.668} & \textbf{0.473} & \textbf{0.980} & \textbf{0.993} \\
            \noalign{\hrule height \heavyrulewidth}
        \end{tabular}
        }

        % \vspace{-8pt}
    \end{minipage}

    % \vspace{-8pt}
\end{table}

\myparagraph{U-Net control methods on DiT.}
Fig.~\ref{fig:fig_7} shows training-free motion-control strategies originally designed for U-Net, such as Noise Warping and SG-I2V, do not transfer well to DiT. Noise Warping produces visually collapsed videos for both object and camera motion control (Fig.~\ref{fig:fig_7}(a)), which is reflected in its poor FID, FVD, and VBench scores in Table~\ref{tab:main}. 
SG-I2V mostly generates static videos for both object and camera control, with only occasional responses to the control signals. As illustrated in Fig.~\ref{fig:fig_7}(b), the penguin’s torso is abnormally stretched along the bounding-box trajectory, suggesting that partial guidance effective in U-Net fails on DiT due to the weaker locality bias in DiT. Consequently, SG-I2V exhibits low FID and FVD in Table~\ref{tab:main}, while producing relatively high VBench scores since the generated videos are largely static with little frame-wise variation. 
Moreover, methods like SG-I2V require ad-hoc search over guidance blocks, which becomes impractical for DiT models with many blocks. In contrast, our method applies query warping across all blocks with negligible overhead and injects guidance only at the final output, eliminating the need for block selection. Further analysis of SG-I2V on DiT is provided in the supplementary material.
% \myparagraph{DiT 위에서의 U-Net control 방법.}
% Fig.~\ref{fig:fig_7}은 U-Net에서 잘 working했던 training-free motion control 전략인 Noise Warping와 SG-I2V가 DiT 위에서는 잘 작동하지 않는다는 것을 보여준다. 우선 noise warping은 object와 camera control 모두에서 시각적으로 완전히 붕괴된 비디오를 produce하며 (Fig.~\ref{fig:fig_7}(a)), 이는 Table~\ref{tab:main}에서 형편없는 FID, FVD, VBench 값에서도 마찬가지다. SG-I2V는 실제로 object와 camera control 모두에서 대부분 아무 움직임이 없는 비디오를 만들었으나, 가끔씩 control에 반응하는 모습이 관찰되었다. Fig.~\ref{fig:fig_7}(b)에서 bounding box trajectory에 따라 펭귄의 몸통이 비정상적으로 늘어난 것을 볼 수 있는데, 이는 U-Net에서 작동했던 부분적인 box 수준 가이던스가 locality에 대한 inductive bias가 낮은 DiT에서는 잘 작동하지 않는 것으로 보인다. 이러한 아티팩트스러운 모션 때문에, Table~\ref{tab:main}에서는 낮은 FID와 FVD를 보였으며, 대부분 완전히 정지된 비디오를 내뱉기 때문에 프레임 별 변화가 없어 상대적으로 높은 VBench 값을 보이는 것을 알 수 있다. 게다가, SG-I2V와 같이, 가이드를 줄 block을 ad-hoc search해야 하는 방법은 블럭 수가 많은 DiT에는 적합하지 않다. 반면에, 우리 방법은 query warping이 연산 오버헤드가 거의 없기 때문에 모든 블럭 내에서 수행하고, 최종 아웃풋에서 guidance를 주기 때문에 이러한 블럭 찾기가 필요 없다. DiT에서의 SG-I2V 대한 further 분석과 discussion은 서플을 참고해라.

% \subsection{Quantitative Results}

% \myparagraph{Object control.}

% \myparagraph{Camera control.}
% \input{sec/tables/ablation_table}

\subsection{Ablation Studies}
\label{sec:ablation}
We conduct extensive ablations to validate each design choice in our framework.
Table~\ref{tab:ablation_A} and Table~\ref{tab:ablation_B} present ablation results on Wan with DL3DV and CogVideoX with VIPSeg, respectively.
Together, results demonstrate that our design is effective across both backbones and motion types.
Additional ablation videos and hyperparameter studies are provided in the supplementary material.
% We evaluate warping variants and latent-optimization configurations to demonstrate that our components yield the strongest performance. 
% Together, the results show that our design is effective across both backbones and motion types, covering camera and object motion control. 
% We conduct extensive ablations to validate each design choice in our framework.
% Table~\ref{tab:ablation_A}와 \ref{tab:ablation_B}에서 각각 Wan, CogVideoX에 대해 We evaluate multiple warping variants and latent-optimization configurations to show that our selected components offer the strongest performance. 두 테이블은 각각 DL3DV, VIPSeg 데이터셋에 대해 구한 것이므로 우리의 design이 두 모델, 그리고 object, camera motion control 둘 다에서 효과가 있음을 체계적으로 보인다. Additional hyperparameter studies are provided in the supplementary material.

\myparagraph{Warping target.}
We first ablate the components involved in query warping. Removing STCD (a), omitting spatial RoPE warping (b), or additionally warping temporal RoPE (c) all degrade performance, confirming that selectively warping only the semantic component is essential for stable control.
We also examine warping other attention components. Warping the keys only (d), warping both queries and keys (e), or warping the attention output (f) all underperform compared to our query-only strategy. The relatively high VBench scores for (e) and Vanilla Wan stem from their tendency to produce largely static videos. This supports our theoretical analysis in Sec.~\ref{sec:why_query}, which suggests that query warping is uniquely suited for concentrating attention on the desired motion region.

\myparagraph{Latent optimization.}
We ablate the components contributing to motion guidance. Using only query-warped denoising without latent optimization (g) yields slightly worse scores yet still retains strong control effects, indicating that query-warped denoising alone provides effective motion guidance (Fig.~\ref{fig:fig_2}(b)). In contrast, performing latent optimization without query-warped denoising (h) results in weaker control.
The performance gap from removing latent optimization is more pronounced for camera control than object control, suggesting that latent optimization can be partially omitted for object control under limited computational budgets.
Finally, removing the phase-consistency constraint (i) degrades performance, highlighting its importance for motion fidelity and stability.
% We ablate the components contributing to motion guidance. Using only query-warped prediction for denoising without latent optimization (g) or performing latent optimization only, without query-warped denoising (h) produces slightly lower scores, yet both retain strong control effects, demonstrating that query-warped denoising alone provides effective motion guidance (Fig.~\ref{fig:fig1}(b)). latent optimization 유무의 성능 차이는 object control보다는 camera control에서 더 두드러지는데, 따라서 연산 상황에 따라 object control에서 latent optimization는 어느정도 생략 가능할 수 있다.
% Finally, adding latent optimization (h) and the phase-consistency constraint (i) yields the strongest performance, confirming that every component of our latent optimization contributes meaningfully to motion fidelity and stability.

\begin{wraptable}{r}{0.57\columnwidth}
\vspace{-30pt}

\centering
\fontsize{7.6}{10}\selectfont
\setlength{\tabcolsep}{1pt}
\renewcommand{\arraystretch}{0.9}

\caption{Effect of optical flow source on camera control.
OF err.: difference between the input and re-estimated flow from the generated video.}
\label{tab:optical_flow}

\begin{tabular}{@{}lccccc@{}}
\toprule
Option
& OF err. $\Downarrow$
% & FVD (×10$^{3}$) $\Downarrow$
& FTD $\Downarrow$
& BC $\Uparrow$
& MS $\Uparrow$ \\
\midrule
% Estimated (VGGT) & 0.710 & 1.294 & 0.251 & 0.931 & 0.938 \\
% Estimated (DAM3) & 0.707 & 1.308 & 0.243 & 0.933 & 0.940 \\
% \rowcolor{MyBlue!70!white}
% Ground-truth (RAFT) & 0.700 & 1.007 & 0.237 & 0.946 & 0.941 \\
Estimated (VGGT) & 0.710 & 0.251 & 0.931 & 0.938 \\
Estimated (DAM 3) & 0.707 & 0.243 & 0.933 & 0.940 \\
\rowcolor{MyBlue!70!white}
Ground-truth (RAFT) & 0.700 & 0.237 & 0.946 & 0.941 \\[-1.5pt]
% \vspace{-0.35\baselineskip}\\[-6.5pt]
\bottomrule
\end{tabular}

% \begin{tabular*}{\columnwidth}{@{\extracolsep{\fill}}lccccc@{}}
% \strut Option
% & \strut OF err. $\Downarrow$
% & \strut FVD (×10$^{3}$) $\Downarrow$
% & \strut FTD $\Downarrow$
% & \strut BC $\Uparrow$
% & \strut MS $\Uparrow$ \\
% \hline
% Depth (VGGT) & 0.710 & 1.294 & 0.251 & 0.931 & 0.938 \\
% Depth (DAM3) & 0.707 & 1.308 & 0.243 & 0.933 & 0.940 \\
% GT (RAFT)    & 0.700 & 1.007 & 0.237 & 0.946 & 0.941 \\
% \end{tabular*}
\vspace{-19pt}
\end{wraptable}
\myparagraph{Effectiveness in practical scenarios.}
To evaluate the practicality of our camera control, we examine whether it remains effective without ground-truth motion signals.
In our experiments, optical flow extracted with RAFT \cite{raft} from the ground-truth video is used as the control signal.
However, in real deployment only a single input image is available, requiring optical flow to be derived from predicted depth and a user-specified camera trajectory.
To simulate this setting, we estimate depth using VGGT \cite{vggt} and Depth Anything 3 (DAM 3) \cite{dam3}, and compute depth-based optical flow using camera poses predicted by VGGT.
Table~\ref{tab:optical_flow} shows that depth-estimated flow from the two models consistently yields results comparable to those obtained with ground-truth flow.
This suggests that our optical-flow-based camera control is practical for real-world use.

\section{Conclusion}
\label{sec:conc}
In this work, we present \MODELNAME, a training-free framework for user-defined control of both object and camera motion in image-to-video DiT models. Our key insight is that query features establish a stable spatial layout early in denoising, allowing motion to be steered by warping queries within the 3D full attention. We introduce STCD to isolate a frame-consistent semantic subspace that can be safely warped without disrupting attention. Query-warped predictions provide strong early-step motion guidance, enabling self-guided latent optimization that improves both motion stability and visual quality. Experiments show that \MODELNAME outperforms existing training-free methods and achieves performance comparable to learning-based approaches while preserving the generalization ability of the pretrained backbone. These results establish query warping as a principled mechanism for controllable DiT-based video generation.

\subsubsection{\ackname} This work was supported in part by the IITP RS-2024-00457882 (AI Research Hub Project), IITP 2020-II201361, NRF RS-2024-003458\allowbreak{}06, and NRF RS-2023-002620.

\bibliographystyle{splncs04}
\bibliography{main}
% \clearpage

% Restore ToC writing for supplementary sections
\let\addcontentsline\savedaddcontentsline

\clearpage

% ===============================================================
% Supplementary Material
% ===============================================================

\appendix
\setcounter{section}{0}
\setcounter{subsection}{0}
\setcounter{figure}{0}
\setcounter{table}{0}
\setcounter{equation}{0}

\renewcommand{\thefigure}{S\arabic{figure}}
\renewcommand{\thetable}{S\arabic{table}}
\renewcommand{\theequation}{S\arabic{equation}}

% \section*{Supplementary Material}
\begin{center}
{\LARGE\bfseries Supplementary Material\par}
\end{center}

\vspace{60pt}

We recommend readers to watch the accompanying \texttt{.mp4} videos included
in the supplementary material, as static frames alone cannot fully convey the
temporal coherence and realism of the generated motion.

\vspace{50pt}

\setcounter{tocdepth}{2}
\renewcommand{\contentsname}{Contents}
\addtocontents{toc}{~\hfill\textbf{Page}\par}

\begingroup
\hypersetup{linkcolor=eccvblue}
\providecommand{\authcount}[1]{}
\tableofcontents
\endgroup

\addtocontents{toc}{\vspace{5pt}}

% \vspace{25pt}
\clearpage

\section{Algorithm}
Overall, Algorithm~\ref{alg:motion_control} interleaves two complementary mechanisms (\ie, self-guided latent optimization and query-warped denoising) to achieve motion control under the user-defined warp $\mathcal{W}$. 
The latent optimization loop in the motion-guided timestep range $\mathcal{T}_{\text{guide}}$ reshapes the early trajectory in latent space toward the desired motion by aligning the standard prediction $\hat{\mathbf{u}}_t$ with the query-warped prediction $\hat{\mathbf{u}}_t^{\text{warp}}$. 
This optimization stage can be optionally skipped depending on the available computational budget, particularly in object control scenarios.
Subsequently, the query-warped predictions provide an explicit motion bias in the solver step, steering the generation trajectory to follow the user-specified motion. 
For timesteps outside the guided range, the model continues with standard predictions $\hat{\mathbf{u}}_t$, allowing the backbone DiT to refine appearance details and stabilize the final reconstruction. 
This lightweight combination enables \MODELNAME to follow user-defined motion cues while preserving both the fidelity and temporal stability of pretrained video DiTs.
% Overall, Algorithm~\ref{alg:motion_control} interleaves two complementary mechanisms (\ie, self-guided latent optimization and query-warped denoising) to achieve motion control under the user-defined warp $\mathcal{W}$. 
% The latent optimization loop in the motion-guided timestep range $\mathcal{T}_{\text{guide}}$ reshapes the early diffusion trajectory in latent space toward the desired motion by aligning the standard prediction $\hat{\mathbf{u}}_t$ with the query-warped prediction $\hat{\mathbf{u}}_t^{\text{warp}}$. 
% Subsequently, the query-warped updates provide an explicit motion bias in the flow-matching solver step, steering the denoising process along the user-specified trajectory. 
% For timesteps outside the guided range, the model continues with standard denoising, allowing the backbone DiT to refine appearance details and stabilize the final reconstruction. 
% This lightweight combination enables \MODELNAME to follow user-defined motion cues while preserving both the fidelity and temporal stability of large DiT-based video generation.

\begin{algorithm}
\caption{Motion-Controlled Sampling with \MODELNAME}
\KwIn{
    Condition image $\mathbf{c}$, User-defined warp $\mathcal{W}$;\\
    \Indp
    Timestep schedule $\{t_0 > t_1 > \dots > t_T\}$\\
    Motion-guided timestep range $\mathcal{T}_{\text{guide}}$\\
    Number of optimization steps $K$, step size $\eta$\\
    \Indm
}
\KwOut{
    Generated video $\hat{\mathbf{v}}$
}

Sample initial latent $\mathbf{x}_{t_0} \sim \mathcal{N}(0, I)$\;

\For{$i \gets 0$ \KwTo $T-1$}{
    $t \gets t_i$, \quad $t_{\text{next}} \gets t_{i+1}$\;

    \If{$t \in \mathcal{T}_{\text{guide}}$}{
        \Comment{Self-guided latent optimization}
        \For{$k \gets 1$ \KwTo $K$}{
            $\hat{\mathbf{u}}_t \gets f_\theta(\mathbf{x}_t, t, \mathbf{c})$\;
            $\hat{\mathbf{u}}_t^{\text{warp}} \gets f_\theta^{\text{warp}}(\mathbf{x}_t, t, \mathbf{c}; \mathcal{W})$\;
            $\mathcal{L}_\text{guide} \gets L_\text{guide}(\hat{\mathbf{u}}_t, \hat{\mathbf{u}}_t^{\text{warp}})$\;
            $\mathbf{x}_t \gets \mathbf{x}_t - \eta \,\nabla_{\mathbf{x}_t} \mathcal{L}_\text{guide}$\;
        }
        \Comment{Query-warped solver step}
        $\hat{\mathbf{u}}_t^{\text{warp}} \gets f_\theta^{\text{warp}}(\mathbf{x}_t, t, \mathbf{c}; \mathcal{W})$\;
        $\mathbf{x}_{t_{\text{next}}} \gets \textsc{SolverStep}(\mathbf{x}_t, \hat{\mathbf{u}}_t^{\text{warp}}, t, t_{\text{next}})$\;
    }
    \Else{
        \Comment{Standard solver step}
        $\hat{\mathbf{u}}_t \gets f_\theta(\mathbf{x}_t, t, \mathbf{c})$\;
        $\mathbf{x}_{t_{\text{next}}} \gets \textsc{SolverStep}(\mathbf{x}_t, \hat{\mathbf{u}}_t, t, t_{\text{next}})$\;
    }
}

Decode $\mathbf{x}_{t_T}$ into video frames $\hat{\mathbf{v}}$\;

\label{alg:motion_control}
\end{algorithm}

\section{Implementation Details}
\label{sec:impl_detail}
For implementation, we referenced the publicly released codebases of
DiTFlow \cite{ditflow} and RoPECraft~\cite{ropecraft}.
All other models are implemented using their official public repositories
for a fair comparison. All experiments are conducted on a single NVIDIA RTX A6000 GPU with 48GB memory. Below, we describe the implementation details of \MODELNAME and the baseline methods.
% For implementation, we DiTFlow와 RoPECraft의 코드를 참고해서 구현했다 for our method, while all other models are implemented based on their publicly released official code for fair comparison. 모든 실험은 a single NVIDIA RTX A6000 48GB GPU에서 진행되었다. 아래에는 \MODELNAME과 baseline method들에 대한 디테일한 구현 사항을 명세한다.
% For implementation, we follow the official source code of \ym{MODEL NAME} for our method, while all other models are implemented based on their publicly released official code for fair comparison.

\subsection{Our Method}
\label{sec:imp_details_ours}
\MODELNAME is implemented on both the Wan 2.2 TI2V-5B~\cite{wan} and
CogVideoX-I2V-5B~\cite{cogvideox} backbones. Query warping is applied to
\emph{all} attention layers, offering a practical advantage over prior
U-Net approaches~\cite{sgi2v,moft} that require manual selection of the
feature block for loss computation. Since our goal is to generate the
video according to the user-defined motion, we use an empty text prompt.
% 우리의 \MODELNAME은 Wan 2.2 TI2V 5B와 CogVideoX-I2V 2B 위에서 각각 구현되었다. query warping은 모든 attention layer에 대해 적용되었으며, 이는 layer 선택하기 위해 노가다가 필요한, layer feature 상에 loss를 주는 과거 UNet에 적용된 방법론들과 비교했을때 장점이다. 또한, 입력 이미지를 주어진 warping에 따라 움직이게 하는 것이 목적이므로 text prompt는 null text ("")으로 주었다.

\myparagraph{\MODELNAME on Wan.}
Wan alternates between (1) full self-attention over all frame tokens and (2) cross-attention from frame-token queries to text-token keys. Because queries in both layers originate from frame tokens, our query warping can be applied uniformly across all layers. For the ``key-warping" ablation in the main paper, warping is applied only to the self-attention layers, where frame tokens serve as keys.
We apply query-warped prediction and latent optimization only during the early timesteps (the first 5 steps out of 50), where the global layout and motion trajectory of the video are determined \cite{motionshop}. At each timestep, we run 5 latent-optimization iterations using the Adam \cite{adam} optimizer, decreasing the learning rate linearly from 0.001 to 0.0005. Sec.~\ref{sec:abl_hyperparameters} provides additional studies on the choice of timesteps for motion guidance. The learning rate schedule was selected empirically by observing the loss dynamics across a variety of samples and choosing the configuration that yielded the most stable results on average.
% Wan은 전체 frame token들 간의 full self-attention layer와 text토큰(key)과 frame 토큰(query) 사이의 cross-attention layer가 분리되어 반복되는 형태이다. 두 레이어 모두 query는 frame 토큰들로 구성되므로 우리의 query warping은 모든 레이어에 적용될 수 있다. 본문의 ablation study에서 Key를 warping했을때는 frame 토큰이 key인 self-attention layer에만 warping이 적용되었다. 우리는 video의 layout 및 global한 모션이 결정되는 초반 스텝(총 50스텝 중 첫 다섯 스텝)에만 query-warped denoising 및 latent optimization을 적용하였다. 또한, 각 타입스텝마다 5번의 optimization을 반복하였으며, learning rate는 각 타입스텝 내에서 optimization step마다 0.001부터 0.0005까지 선형적으로 감소시켰다. 우리는 Sec. \ref{sec:additional_exp}에서 query warping을 적용하는 타임스텝과 optimization 반복횟수에 대한 study를 수행한다. learning rate는 loss 값이 변화하는 양상을 데이터 샘플마다 관찰하고 평균적으로 가장 좋은 세팅을 경험적으로 설정한 것이다. 

\myparagraph{\MODELNAME on CogVideoX.}
CogVideoX performs self-attention jointly over concatenated text and frame tokens. We therefore apply warping only to the query tokens corresponding to video frames. As with Wan, we use query-warped prediction and latent optimization only for the first 5 timesteps, running 5 optimization iterations per timestep with Adam optimizer. The learning rate is linearly annealed from 0.02 to 0.005 based on empirical tuning.
% CogVideoX는 text 토큰과 frame 토큰이 concat된 전체 토큰에 대해서 self-attention을 수행한다. 우리는 query의 frame 토큰에만 warping을 적용하였다. Wan에서와 유사하게, 초반 다섯 timestep에만 query-warped denoising과 latent optimization을 수행하였으며, 각 타입스텝마다 5번의 optimization을 반복하였다. learning rate는 경험적으로 ?.???부터 ?.???까지 선형적으로 감소시켰다.

\myparagraph{Applying the Warps.}
In Wan 2.2, the latent patches are downsampled by a factor of 32 in space and 4 in time relative to the original video. In CogVideoX, the spatial downsampling is 16× and the temporal downsampling is 4×. Accordingly, we downsample the user-specified mask warps (for object control) or optical flow fields (for camera control) to the latent resolution before applying query warping.
For the temporal dimension, we apply mask warping with a stride of 4, since the user-defined warp is specified as a single mapping from frame 1 to frame $n$, rather than as per-frame displacements. Optical flow, which encodes per-frame displacements, is accumulated across 4 latent frames to match the temporal compression factor.
% \myparagraph{warp 적용.} latent diffusion model인 Wan 2.2는 latent patch의 spatial 해상도가 원본 비디오보다 32배, temporal 해상도가 원본 비디오보다 4배 압축되어 있다. CogVideoX는 latent patch의 spatial 해상도가 원본 비디오보다 16배, temporal 해상도가 원본 비디오보다 4배 압축되어 있다. 따라서 우리는 원본 비디오 해상도에서 정의된 mask warping (object control)이나 optical flow (camera control)을 각각의 latent 패치 해상도로 다운샘플 하여서 적용했으며, temporal 축으로는 mask warping은 stride 4로 적용하고 (첫프레임에서 n프레임으로의 warping이 정의돼 있으므로), optical flow는 4 프레임의 warping을 combine해서 (프레임 n에서 다음 프레임 사이의 displacement이므로) 적용하였다.

\subsection{Baseline Methods}
% \ym{DL3DV 는 reference 달았는데 모델 부분도 달아야되는건지. 만약에 달아야되면 모델 부분에 reference 삭제하는게 나을수도}

\myparagraph{SG-I2V.}
We used the original model without modifications, adapting only the dataset input format to match the requirements of SG-I2V \cite{sgi2v}. 
Implementation details and additional analysis of SG-I2V on DiT are provided in Sec.~\ref{sec:sgi2v_on_dit}.
Bounding boxes and trajectories across all frames were provided as motion prompts. Details of bounding box and trajectory acquisition are provided in Sec.~\ref{sec:datasets}.
% Each bounding box was defined by treating the trajectory point as its center and constructing a square inscribed in the largest circle fully contained within the object region. 
Videos were generated at the default resolution of 576 $\times$ 1024. The number of generated frames matched the reference videos; however, due to GPU memory limitations, sequences longer than 25 frames were truncated to 25. 

\myparagraph{FreeTraj.}
We applied FreeTraj \cite{freetraj} using the official I2V setting provided in its official repository, employing VideoCrafter-I2V~\cite{videocrafter1} as the backbone model. During generation, we fixed the resolution to $512 \times 512$ and set the number of frames equal to that of the reference video for each sample. For motion control, we guided object motion using the same bounding-box and trajectory specifications as SG-I2V. Since it does not support multiple bounding boxes as input, we selected a single representative object trajectory for control in each scene.
% \ym{얘네 bbox 기준. 코드에서 I2V 말해서 그거 따름.}
% \ym{bbox 여러개 못받길래 하나의 대표적인 움직임만 했다라는걸 적어야하나? 숨겨야하나?} \chu{적어도됨}
% \begin{itemize}
%     \item Resolution: $480 \times 832$
%     \item Frame num: reference video 와 동일하게.
%     \item Backbone: VideoCrafter-I2V
%     \item How to control: Warping \chu{입력 방법은 SG-I2V랑 동일하니까 SG-I2V 밑으로 옮겨서 똑같이 했다 이렇게 해도 될듯.}
% \end{itemize}
% 우리는 FreeTraj를 적용하기 위해 official repository 에 있는 I2V setting 으로 backbone 을 VideoCrafter-i2v 를 사용했다. 비디오를 생성할 떄 512 x 512의 resolution 을 따르고 비디오의 생성 프레임 개수는 dataset의 reference video 와 동일하게 셋팅하였다. 그리고 motion control의 방식은 SG-I2V 방식과 동일하게 사용하였다.

\myparagraph{MOFT.}
MOFT~\cite{moft} employs its coarse motion control module together with DIFT~\cite{dift} for point-level control. Following the original implementation, we applied Content Correlation Removal to the features extracted after upper block 1. The input data format was identical to that used for SG-I2V. Videos were generated at a fixed resolution of 576 $\times$ 1024. The number of generated frames matched the input sequence length, but sequences longer than 25 frames were truncated to 25 due to memory constraints.

\myparagraph{MotionClone.}
We follow the official MotionClone~\cite{motionclone} implementation and only replace its AnimateDiff~\cite{animatediff} backbone from T2V to I2V. We generate videos at the default resolution of $512\times512$ and default length of 16 frames. We attempted to increase the number of frames up to the limit of our GPU memory to enable evaluation on longer sequences, but found that MotionClone produced severely degraded videos whenever the frame length differed from 16, so we retained the default setting. Since MotionClone is originally a motion-transfer method that takes a motion reference video and generates a new video with similar motion, we feed the ground-truth video as the reference to perform motion control. This gives MotionClone a highly favorable evaluation setting, as it observes the ground-truth motion.
% 우리는 motionclone의 공식 구현을 그대로 유지한 채 backbone인 AnimateDiff만 T2V에서 I2V로 바꿔서 사용했다. 우리는 MotionClone의 디폴트 resolution인 512*512와 디폴트 프레임 길이인 16으로 비디오를 생성했다. 우리는 더욱 긴 프레임 길이로 평가하기 위해 프레임 길이를 우리 GPU 메모리의 한계까지 늘려서 생성해보았지만, 프레임 길이가 16이 아니면 완전히 망가진 비디오를 만들었기 때문에 이는 어쩔 수 없는 선택이었다. MotionClone은 원래 motion reference video를 입력받아 그와 유사한 모션을 가진 비디오를 만드는 motion transfer method이기 때문에, 우리는 ground truth 비디오를 reference video로 넣어 주어 motion control을 하였다. 이는 ground truth video를 입력받기 때문에 매우 유리한 세팅이다.
% We implemented this method based on the official codebase, using its default configuration. Videos were generated at a resolution $512 \times 512$, and due to memory constraints on a single NVIDIA A6000 GPU, the maximum generation length was limited to 16 frames. Motion control was achieved by providing the ground‐truth video as the motion reference, enabling the model to reproduce its motion patterns.
% \ym{A6000 으로 해서 최대 생성 길이 16 frame 까지 안되는데, 그거까지 적기.}
% \begin{itemize}
%     \item Resolution: 512 $\times$ 512
%     \item Frame num: 16 (최대 생성 길이)
%     \item How to control: Video 로 motion 줌. the ground-truth video as a motion reference video for the motion transfer method.

% \end{itemize}

\myparagraph{Go-with-the-flow.}
Go-with-the-flow (GWTF)~\cite{gowiththeflow} is a motion-controllable video generation model finetuned on the CogVideoX-5B-I2V~\cite{cogvideox} backbone. We follow its default configuration and generate videos at a resolution of $480\times 720$ with 49 frames, applying warping to control motion. Similar to MotionClone, GWTF produces severely degraded outputs if the number of frames differs from 49. Therefore, when the target video length is shorter than 49 frames (as in VIPSeg), we extend the input optical flow to match 49 frames.
A naive strategy of simply repeating the last-frame optical flow also leads to failure, so we instead ``ping-pong" the optical flow sequence in time. Concretely, for a sequence of flows over frames $1,2,3,4$, we extend it as
$1,2,3,4,3,2,1,2,3,4,3\ldots$
by repeatedly reversing the direction at the endpoints. This yields temporally continuous yet non-identical optical flow across frames, allowing us to reach 49 frames without collapsing the generated video.

\myparagraph{Noise warping on DiT.}
FreeTraj~\cite{freetraj} showed that in U-Net–based video generation models, simply warping the input noise can induce corresponding motion. To examine whether this property also holds for DiT-based models, we apply the same noise warping to the input latent of Wan. While FreeTraj further improves control stability through attention masking, this mechanism cannot be directly applied to DiT architectures and is also infeasible in scenarios where foreground and background cannot be separated, such as camera motion control. Therefore, we evaluate whether pure noise warping alone achieves results comparable to the preliminary findings reported in FreeTraj. For warping, we adopt the recent algorithm proposed in GWTF~\cite{gowiththeflow}, which preserves Gaussianity during the warping process.
Following FreeTraj, we resample $75\%$ of the high-frequency noise after warping.

% U-Net 기반 video 생성 모델에서는 단순히 입력 noise를 warping만 해도 그에 따른 모션이 유도된다는 FreeTraj~\cite{freetraj}의 발견에 따라, 우리는 DiT에서도 똑같이 먹히는지 확인하기 위해 noise를 warping해서 DiT 기반 모델에 넣어보았다. FreeTraj에서는 추가적인 attention masking 기법을 통해 컨트롤 안정성을 높였지만, 이는 DiT에서는 단순하게 적용이 불가능하기 때문에, 순수 noise warping만 했을때 FreeTraj에서 나온 priliminary 결과와 유사한 수준으로 나오는지 확인하는 것이 목적이었다. noise warping은 gaussianity를 안정적으로 유지하면서 noise를 warping하는 GWTF의 최신 샘플링 알고리즘을 이용하였다. 

% GWTF was implemented using a fine-tuned CogVideoX-5B-I2V~\cite{cogvideox} model, generating videos at a resolution of $480 \times 720$ with 49 frames and applying warping for motion control. 
% In cases where the original video contained fewer than 49 frames, we extended the sequence using temporal back-and-forth looping to ensure a consistent length for quantitative evaluation.
% When the original video contained fewer than 49 frames, we extended it using temporal back-and-forth looping to maintain a consistent sequence length. 
% This avoids the lack of motion cues that would arise from static frame repetition, which is critical since this method relies on optical-flow-based warping for motion control.
% This avoids the motion discontinuity that would occur if the final frame were simply repeated or padded.
% 이는 

% \subsection{}

% \input{supp/figs/timestep}
% \input{supp/figs/ksem}
% \input{supp/figs/timestep}
% \input{supp/figs/ksem}
\begin{figure*}[t]
    \centering
    \includegraphics[width=\textwidth]{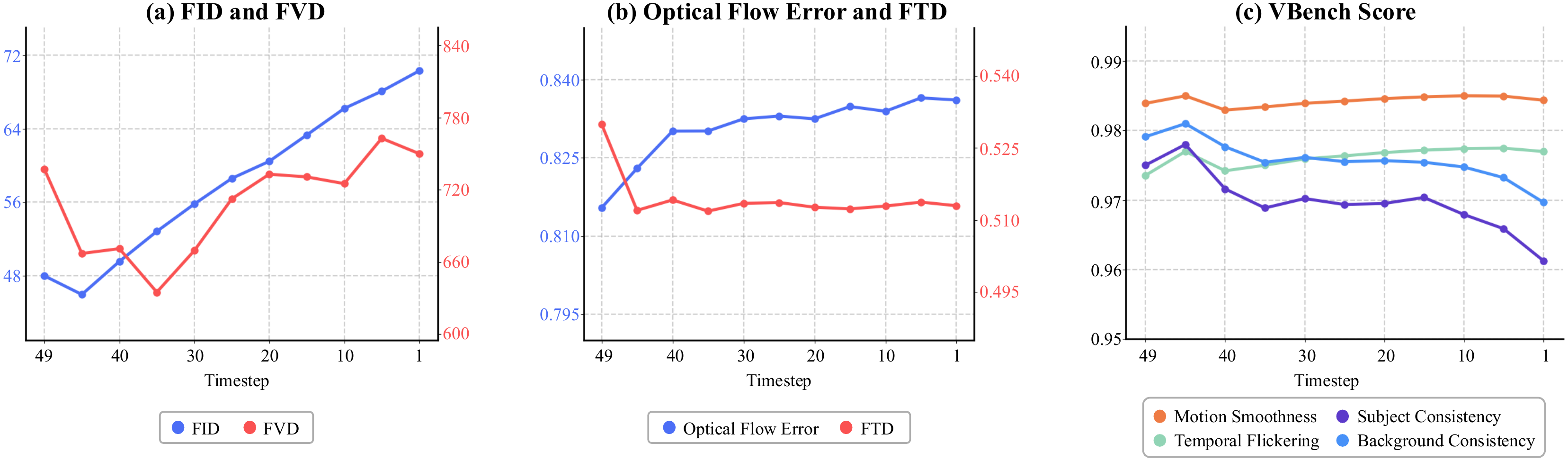}
    % \vspace{-19pt}  % 너무 세게 줄이면 글자랑 겹칠 수 있음. 필요하면 -4pt 정도만.
    \caption{
    Ablation on the timestep range for applying motion guidance. We vary how long motion guidance is applied starting from the first guidance step $t_{50}$. Overall, FID, FVD, and optical flow error become worse as the range is extended, while FTD shows no clear trend and VBench achieves its best value around \(t=45\). Based on this trade-off, we stop guidance at \(t=45\), corresponding to the first five denoising steps, where VBench is maximized while FID, FVD, and optical flow error remain favorable.
    % Ablation on motion guide를 주는 timestep range. 첫 타입스텝인 $t_{50}$부터 몇 번째 타입스텝까지 모션 가이드를 주는지이다. 전반적으로 FID, FVD, optical flow error는 timestep range를 넓힐수록 점점 더 안좋아지는 경향을 보인다. FTD는 딱히 경향성을 보이지 않았으며, VBench는 45스텝 부근에서 가장 좋았다. 우리는 FID, FVD, optical flow error가 좋은 초반 스텝들에서 VBench가 가장 좋은 45스텝을 선택했다.
    % Ablation on semantic-temporal channel decomposition에서의 k_sem과 k_tmp. 우리는 단순성을 위해 k_sem = k_tmp로 놓고 실험하였다. 전반적으로 뚜렷한 경향성을 보이지 않으나, FID, FVD, VBench Score는 k_sem = k_tmp가 3일때 가장 좋았고, optical flow error와 FTD는 들쭉날쭉했다.
    }
    \label{fig:timestep}
\end{figure*}
\begin{figure*}[t]
    \centering
    \includegraphics[width=\textwidth]{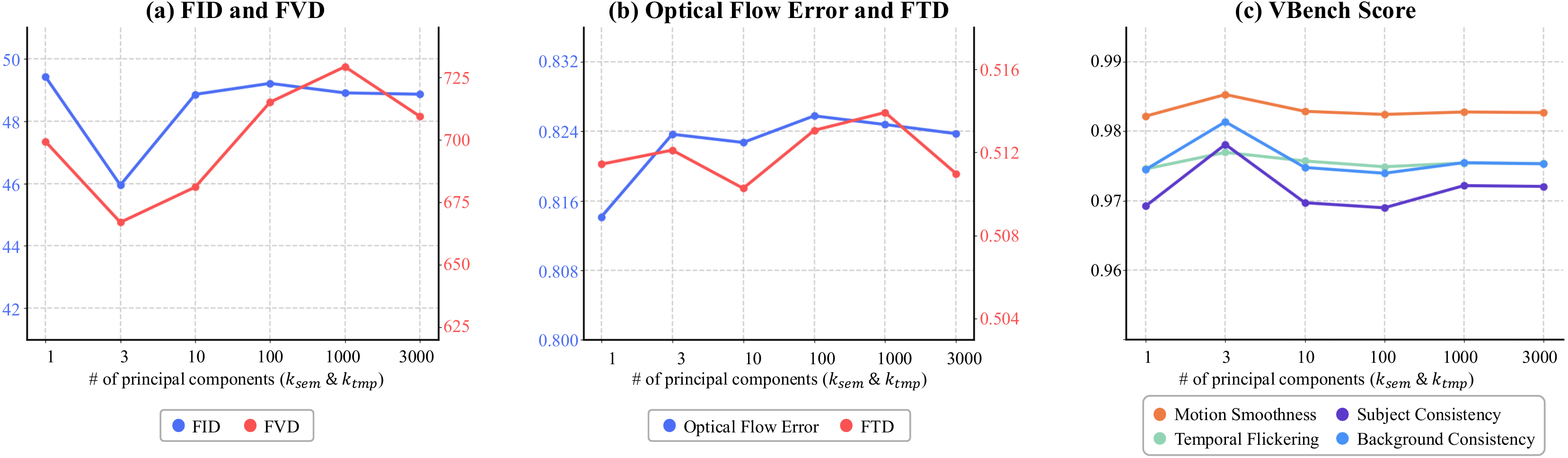}
    % \vspace{-19pt}  % 너무 세게 줄이면 글자랑 겹칠 수 있음. 필요하면 -4pt 정도만.
    \caption{
    Ablation on the number of principal components $k_{\text{sem}}$ and $k_{\text{tmp}}$ in the semantic–temporal channel decomposition. For simplicity, we set $k_{\text{sem}} = k_{\text{tmp}}$ in all experiments. Although the overall trends are not strictly monotonic, FID, FVD, and VBench achieve their best values at $k_{\text{sem}} = k_{\text{tmp}} = 3$, while optical flow error and FTD fluctuate across settings.
    % Ablation on semantic-temporal channel decomposition에서의 k_sem과 k_tmp. 우리는 단순성을 위해 k_sem = k_tmp로 놓고 실험하였다. 전반적으로 뚜렷한 경향성을 보이지 않으나, FID, FVD, VBench Score는 k_sem = k_tmp가 3일때 가장 좋았고, optical flow error와 FTD는 들쭉날쭉했다.
    }
    \label{fig:ksem}
\end{figure*}

\section{Additional Analysis and Experiments}
\label{sec:additional_exp}

\subsection{Hyperparameters}
\label{sec:abl_hyperparameters}
We conduct an ablation study on two key hyperparameters required by \MODELNAME, in order to justify the choices used in our experiments and to better understand their impact on both video quality and motion controllability.
% 우리는 \MODELNAME 구현에 필요한 두 가지 핵심 hyperparameters에 대한 탐색? 스터디를 수행하여 우리의 선택이 best 였음을 증명한다?(이거보다는 더 온건한 톤으로).

\myparagraph{Timesteps for Motion Guidance.}
We first study the range of timesteps at which we apply the motion guidance, that is, query-warped denoising and latent optimization as introduced in Secs.~\ref{sec:query_warping} and~\ref{sec:motion_guidance} of the main paper. Starting from the first timestep $t_{50}$, we vary the endpoint in increments of five timesteps and evaluate how long we should continue applying the guidance.

% Fig.~\ref{fig:timestep} shows that applying motion guidance only during the very early part of the diffusion trajectory (approximately the first five timesteps) provides a favorable trade-off between video quality (FID, FVD, VBench) and motion-control performance (FTD, optical flow error). In particular, as analyzed in Sec.~\ref{sec:sampling_time}, latent optimization introduces a non-negligible computational overhead, so the fact that strong performance can already be achieved by restricting it to the earliest timesteps is especially beneficial in practice.

Fig.~\ref{fig:timestep} shows that extending the timestep range tends to degrade overall video quality metrics (FID, FVD, VBench) as well as optical flow consistency, while the improvement in FTD saturates around \(t=45\). Hence, we select the first five timesteps as the motion-guided range, which sufficiently enhances FTD while maintaining visual quality.
Moreover, as analyzed in Sec.~\ref{sec:sampling_time}, latent optimization introduces a non-negligible computational overhead, so the fact that strong performance can already be achieved by restricting it to the earliest timesteps is especially beneficial in practice.

% \myparagraph{모션을 가이드하는 타입스텝}
% 우리는 main paper의 Sec. 3.2와 Sec. 3.3에서 소개된 query-warped denoising과 latent optimization을 할 타입스텝 범위에 대한 스터디를 진행했다. 첫 타입스텝인 50부터 몇 타입스텝까지 이러한 motion guide를 주는게 최적인지 5 스텝 단위로 실험했다. Fig.~\ref{fig:timestep}은 우리가 선택한 45스텝 부근 (초반 5스텝)에서 비디오 퀄리티(FID, FVD, VBench)와 모션 컨트롤 성능 (FTD, Optical flow error) 사이의 최적의 tradeoff를 보여준다. 특히, Sec.~\ref{sec:sampling_time}에서 분석한 것처럼 latent optimization의 연산 오버헤드가 상당한데 극 초반 스텝만 했을때 성능이 좋다는 것은 참 좋은 일이다.

\myparagraph{Number of Principal Components.}
We also study the number of principal components, $k_{\text{sem}}$ and $k_{\text{tmp}}$, used to compute the channel saliency scores in the Semantic-Temporal Channel Decomposition (STCD) introduced in Sec.~\ref{sec:channel_selection} of the main paper. For simplicity, we set $k_{\text{sem}} = k_{\text{tmp}}$ and vary this shared value over $\{1, 3, 10, 100, 1000, 3000\}$.

As shown in Fig.~\ref{fig:ksem}, setting $k_{\text{sem}} = k_{\text{tmp}} = 3$ yields slightly worse motion-control performance (in terms of FTD and optical flow error) compared to $k_{\text{sem}} = k_{\text{tmp}} = 10$, but leads to noticeably better scores on video quality metrics such as FID, FVD, and VBench. Given this trade-off, we adopt $k_{\text{sem}} = k_{\text{tmp}} = 3$ in our main experiments.
% main paper의 Sec. 3.1에서 소개된 Semantic-Temporal Channel Decomposition의 channel saliency score를 구하는데 사용될 top principal components의 수인 k_sem과 k_tmp에 대한 스터디이다. 우리는 k_sem과 k_tmp는 같은 값으로 단순화하고 1, 3, 10, 100, 1000, 3000으로 늘려보았다. Fig.~\ref{fig:ksem}에서 k_sem = k_tmp = 3에서 10일 때보다 motion control 성능 (FTD와 optical flow error)는 살짝 안좋았지만, 나머지 비디오 퀄리티 관련 성능 (FID, FVD, VBench)은 비교적 큰 폭으로 좋았기 때문에 우리는 3을 선택하였다.

% \input{supp/figs/Hole_filling}                               

\subsection{Hole-Filling Strategy}
\label{sec:hole_filling}
As described in Sec.~\ref{sec:query_warping} of the main paper, we investigate how different strategies for handling hole regions, which arise from user-defined warping, affect the final video quality.
% Our goal is to minimize visual artifacts while ensuring that the intended motion is faithfully reflected in the generated video.
In our pipeline, we apply the user-specified warp to the query tokens of the first frame and then paste the warped content onto the queries of subsequent frames. We adopt different strategies depending on the type of warp.
% For object control, warping is defined by a user-provided mask, whereas for camera control, warping is guided by optical flow.
% Sec. 3.2에 소개된 것과 같이, 우리는 비디오의 아티팩트를 최소화하고 원하는 모션이 잘 반영되게 하기 위해서 user-defined warping 시 발생하는 hole 영역을 처리하는 전략에 따른 결과의 차이를 보여준다. 우리는 첫프레임 쿼리에 입력 와핑을 적용하여 이후 프레임에 붙여넣는다. 이 때 mask을 warping하는 것으로 정의되는 object control과 optical flow 기반의 warping으로 정의되는 카메라 컨트롤에 대해서 각각 다른 전략이 사용된다.

\begin{figure*}[t]
    \centering
    \includegraphics[width=\textwidth]{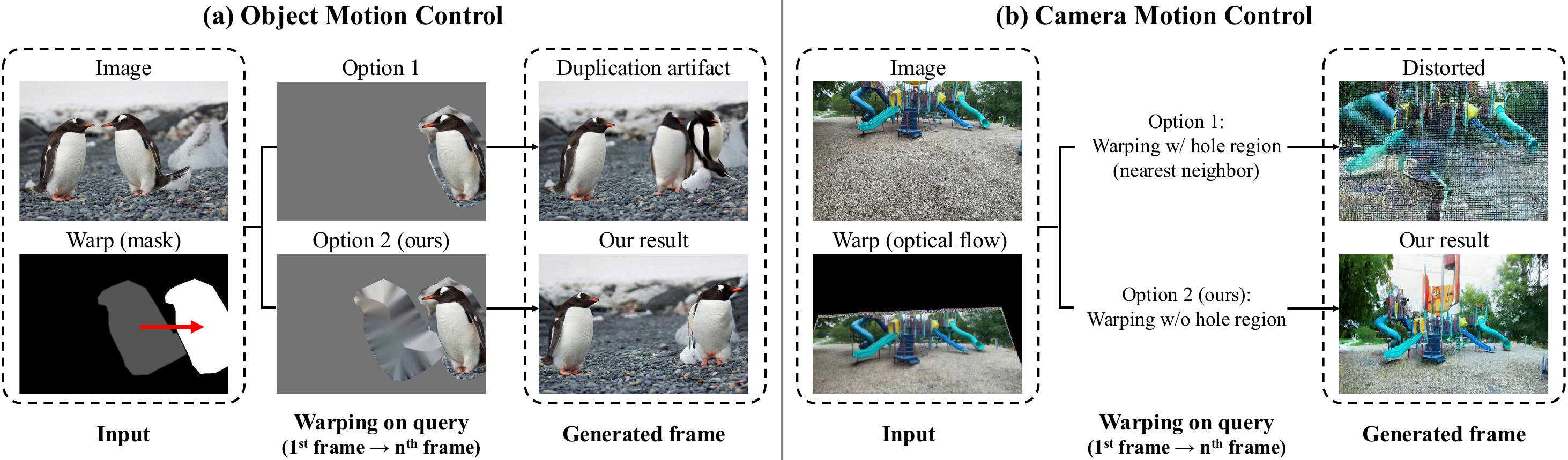}
    % \vspace{-19pt}  % 너무 세게 줄이면 글자랑 겹칠 수 있음. 필요하면 -4pt 정도만.
    \caption{
    Hole-filling strategies in query warping and their impact on video generation. We use different hole-filling schemes for mask-based warping and optical-flow-based warping. With alternative options, the generated videos exhibit noticeable artifacts, whereas our strategy produces videos that more faithfully follow the desired motion.
    % 우리의 hole-filling 전략과 그에 따른 비디오 생성 결과. 우리는 mask warping과 optical flow warping 시 다른 hole-filling 전략을 사용하며, 다른 옵션으로 했을때는 결과 비디오가 아티팩트가 생기는 반면, 우리의 전략을 사용했을때는 원하는 모션의 비디오가 잘 만들어진다.
    }
    \label{fig:hole_filling}
\end{figure*}

\myparagraph{Warping with a Mask (Object Control).}
Fig.~\ref{fig:hole_filling}(a) compares two options for pasting the warped first-frame query onto the $n$-th frame:
\begin{itemize}
    \item \textbf{Option 1:} Paste only the warped mask region from the first-frame query onto the $n$-th frame query.
    \item \textbf{Option 2 (ours):} In addition to pasting the warped mask region, we also fill the hole left by removing the masked object region in the first-frame query with background tokens from the first frame, and paste this hole-filled background together onto the $n$-th frame query.
\end{itemize}
By filling the hole regions with background tokens from the first frame and pasting them along with the warped object, we encourage the attention in the hole areas of later-frame queries to focus on the surrounding background keys of the first frame. This makes the object appear only at its new, warped location and prevents it from remaining at its original position. Fig.~\ref{fig:hole_filling}(a) illustrates this difference in practice. When using Option 1, the object remains visible both at its original position and at the new location, which results in a duplication artifact. In contrast, when using Option 2, only the moving object at its new location is preserved.

% \myparagraph{Warping with a mask (object control).}
% Fig.~\ref{}는 두가지 옵션을 비교한다: (1) 첫프레임 쿼리의 mask 영역을 warping한 것만 n번째 프레임 쿼리에 붙여 넣는다. (2) 첫프레임 쿼리의 mask 영역을 warping한 것과, 첫번째 프레임 쿼리의 마스크 영역을 제거한 hole 영역을 첫프레임 쿼리의 nearest neighbor background 토큰으로 메운 것을 함께 붙여넣는다. 두번째 옵션이 우리의 전략이며, 이후 프레임 쿼리 hole 영역의 어텐션을 첫프레임 key의 주변 background 영역으로 강제함으로써 object가 이동한 영역에만 존재하고 원래 있던 영역에서는 사라지게 한다. Fig.~\ref{}은 첫 번째 옵션을 사용한 경우 object가 원래 자리에도 그대로 남아있는 복제 아티팩트가 나타난 것을 보여주며, 반면에 두 번째 옵션을 사용한 경우 이동하는 오브젝트만 남아있는 것을 볼 수 있다.

\myparagraph{Warping with Optical Flow (Camera Control).}
Similarly, Fig.~\ref{fig:hole_filling}(b) compares two warping options for handling the grid holes and newly exposed empty regions that arise from optical-flow-based warping:
\begin{itemize}
    \item \textbf{Option 1:} Apply backward warping to the first-frame queries and fill all holes and empty regions using a nearest-neighbor scheme, then paste the fully warped and filled queries onto the $n$-th frame queries.
    \item \textbf{Option 2 (ours):} When forward warping is applied, do not paste the values corresponding to grid holes or newly exposed empty regions. Instead, keep the original $n$-th frame query values in those regions.
\end{itemize}
Option 1 fills the hole regions with residual content that behaves like ghosting, which prevents the model from producing the desired motion and leads to stretched structures and severe artifacts, ultimately degrading the video. In contrast, our Option 2 leaves these empty regions to be completed by the model's generative capability, resulting in visually natural content that better preserves the intended camera motion.
% 마찬가지로, Fig.~\ref{}(b)에서 optical flow warping으로 인해 생기는 grid hole과 화면에 없던 빈 영역들을 처리하는 두 가지 warping 옵션에 대한 결과를 비교하고 있다.:
% \begin{itemize}
%     \item \textbf{Option 1:} 첫 프레임 쿼리에서 backward warping 및 nearest neighbor로 hole과 빈 영역을 모두 채운 후 이를 n번째 프레임 쿼리으로 붙여넣는다.
%     \item \textbf{Option 2 (ours):} forward warping으로 생긴 grid hole과 빈 영역에 대해서는 붙여넣지 않고, 원래 n번째 프레임 쿼리 값을 유지한다.
% \end{itemize}
% option 1은 hole 부분이 잔상?으로 채워지기 때문에 결국 원하는 모션을 만들지 못하고 stretch되고 심한 아티팩트와 함께 망가진 비디오를 생성한다. 반면, 우리의 option 2는 빈 영역들을 모델의 생성 능력으로 자연스럽게 채운 것을 볼 수 있다.

% \input{supp/tables/dit_table}

\subsection{SG-I2V on DiT}
\label{sec:sgi2v_on_dit}
As mentioned in Sec.~\ref{sec:intro} of the main paper, before developing our method, we first attempted to adapt a representative and promising training-free motion control approach originally designed for U-Net backbones to a DiT backbone in as faithful a manner as possible. The key idea is to optimize the input latent such that, at a specific block, the feature map within the input bounding box follows the given trajectory, making the bounding box region in the first frame and that in the $n$-th frame match. This optimization strategy is used in SG-I2V~\cite{sgi2v} and MOFT~\cite{moft}, and we followed the more recent variant, SG-I2V.

\begin{figure*}[t]
    \centering
    \includegraphics[width=\textwidth]{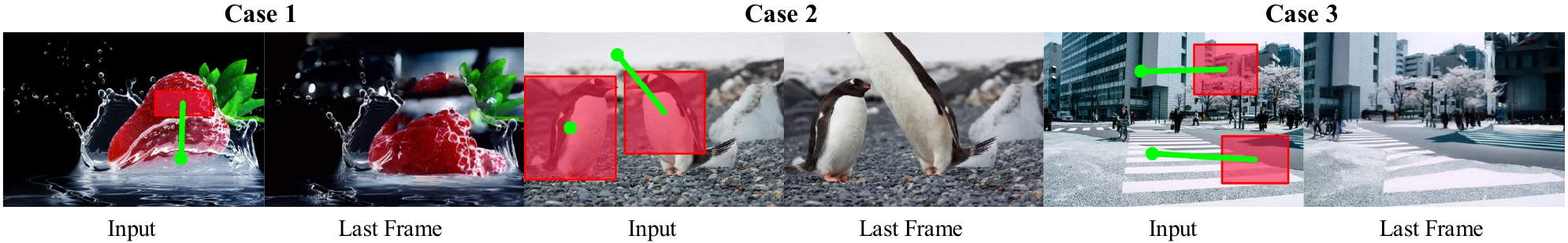}
    % \vspace{-19pt}  % 너무 세게 줄이면 글자랑 겹칠 수 있음. 필요하면 -4pt 정도만.
    \caption{
    Results of implementing SG-I2V \cite{sgi2v} on a DiT backbone (Wan 2.2 TI2V-5B). In most cases, the model produces almost no motion, and even in cases where the object appears to follow the bounding box trajectory, the video becomes unnatural, with only the region inside the bounding box moving while the surrounding context remains static. These examples are originally successful motion-control cases in SG-I2V, which highlights that an SG-I2V-style bounding-box–restricted feature map loss is not well suited for DiT architectures.
    % SG-I2V를 DiT 위에서 구현한 결과. 대부분의 케이스에서 아예 움직임이 없었고, 이와 같이 bounding box trajectory에 따른 움직임을 보인 케이스들은 모두 맥락을 이해하지 못하고 bounding box 부분만 이동하는 부자연스러운 비디오를 생성하였다. 참고로 이 케이스들은 원래 SG-I2V에서 성공적인 motion control을 하는 케이스들이며, 이는 DiT에 SG-I2V와 같이 bounding box 기반의 부분적인 feature map loss를 주는 방식이 적합하지 않음을 보여준다.
    }
    \label{fig:fig_sgi2v}
\end{figure*}

\myparagraph{Implementation.}
SG-I2V uses a \emph{modified attention output} with improved frame consistency as the feature for loss computation, in which the key and value of the $n$-th frame, $K_n$ and $V_n$, are replaced with those of the first frame, $K_1$ and $V_1$. However, due to the 3D full-attention mechanism in DiT, this replacement strategy cannot be directly applied, so we instead use the standard attention output as the target feature. In addition, following SG-I2V, we apply Fast Fourier Transformation (FFT) to the optimized latent $z_t^{\ast}$ and replace its high-frequency components with those of the original latent $z_t$. We implement this approach on the Wan TI2V-5B backbone and use the attention output from block 9, which empirically works best among all blocks, for the loss computation.
% For optimization, we use the Adam optimizer, linearly decaying the learning rate from 0.005 to 0.001 over the inner optimization iterations at each time step.

\myparagraph{Results.}
In our experiments, most cases produced videos with virtually no motion. In a few cases, the object appeared to move along the bounding box trajectory, but as shown in Fig.~\ref{fig:fig_sgi2v}, these were not true successes. The model failed to understand the context of the motion and generated unnatural videos in which only the region inside the bounding box moved while the surrounding scene remained static. 
All of these examples are cases where the original SG-I2V, which uses a U-Net architecture, successfully controlled motion.

We hypothesize that this discrepancy arises from the architectural difference between U-Net and DiT. U-Net is composed of convolutional blocks and therefore has a strong inductive bias toward locality, which makes it more amenable to such localized motion guidance. In contrast, DiT does not have this inherent locality bias, so optimizing with a loss applied only to a partial, bounding-box-defined region of the feature map does not easily lead to coherent, scene-level motion. Furthermore, except for block 9, using other blocks either resulted in an even more pronounced lack of motion or caused the video to collapse entirely, which suggests that applying a loss to a specific feature map is not well suited to DiT in the first place.

Consistent with these observations, Table~\ref{tab:main} in the main paper shows that, in the object motion control setting, FID becomes abnormally high, indicating that the scene is severely degraded, whereas in the camera control setting, FID and FVD are poor but VBench scores remain high, implying that the generated videos are static. A more detailed interpretation of these metrics is provided in Sec.~\ref{discuss:evaluation}.

\begin{table}[t]
\centering
\caption{
Efficiency--controllability trade-off of \MODELNAME and training-free baselines.
Runtime and peak memory are measured for 49-frame generation.
QW denotes query warping; SG-I2V on DiT is implemented nearly identically to MOFT.
}
\resizebox{1.0\linewidth}{!}{
\begin{tabular}{llcccc}
\toprule
Method & Backbone & Latent opt. & Runtime (sec) $\downarrow$ & Peak memory (GB) $\downarrow$ & FTD $\downarrow$ \\
\midrule
No control & Wan & -- & 120.38 & 17.59 & 0.367 \\
Noise warping & Wan & X & 125.11 & 17.59 & 0.455 \\
SG-I2V ($\approx$ MOFT) & Wan & O & 297.88 & 39.51 & 0.438 \\
\rowcolor{MyBlue!70!white}
\MODELNAME (QW only) & Wan & X & 132.62 & 17.61 & 0.274 \\
\rowcolor{MyBlue!70!white}
\MODELNAME (QW + opt.) & Wan & O & 314.34 & 40.17 & 0.237 \\
\specialrule{\lightrulewidth}{0pt}{\belowrulesep}
No control & CogVideoX & -- & 452.01 & 23.30 & 0.537 \\
\rowcolor{MyBlue!70!white}
\MODELNAME (QW only) & CogVideoX & X & 474.47 & 23.78 & 0.501 \\
\rowcolor{MyBlue!70!white}
\MODELNAME (QW + opt.) & CogVideoX & O & 1092.66 & 47.59 & 0.473 \\
\specialrule{\heavyrulewidth}{0pt}{0pt}
\end{tabular}
}
\label{tab:suppl_sampling_time}
\end{table}

% \begin{wraptable}{r}{0.50\linewidth}
% \vspace{-35pt}
% \input{supp/tables/sampling_time_inline}
% \vspace{-20pt}
% \end{wraptable}

% \subsection{Sampling Time}
\subsection{Efficiency--Controllability Trade-off}
\label{sec:sampling_time}
We analyze the efficiency--controllability trade-off of \MODELNAME by jointly reporting runtime, peak memory, and motion controllability. All runtime and memory measurements were conducted for 49-frame generation on the same GPU, and each runtime is averaged over 10 runs. For query warping, we report the more expensive optical-flow-based setting. Since the two backbones are evaluated on different motion-control settings in our experiments, FTD is reported on DL3DV camera control for Wan and VIPSeg object control for CogVideoX, following our evaluation protocol.

As shown in Table~\ref{tab:suppl_sampling_time}, query warping adds only a small overhead to the vanilla model. On Wan, \MODELNAME without latent optimization increases runtime by about 12 seconds and peak memory by only 0.02 GB, while achieving substantially better FTD than other training-free baselines adapted to DiT. On CogVideoX, query warping similarly adds only about 22 seconds and 0.48 GB relative to the much larger vanilla sampling cost, while improving FTD from 0.537 to 0.501. These results indicate that the query-warping component itself is lightweight and already provides effective motion control.

Latent optimization further improves controllability, reducing FTD from 0.274 to 0.237 on Wan and from 0.501 to 0.473 on CogVideoX. This comes with non-negligible runtime and memory overhead, but the cost is comparable to strong optimization-based baselines such as SG-I2V. Thus, latent optimization can be enabled when stronger controllability is required, while query warping alone provides a more efficient alternative. For longer generation on Wan, increasing the number of frames from 49 to 97 keeps the query-warping time overhead around \(+13\) seconds, while the relative optimization overhead decreases from \(2.6\times\) to \(2.2\times\).

\subsection{Human Evaluation}
\label{sec:human_eval}

To complement automatic metrics and qualitative comparisons, we conducted a human evaluation on 19 motion-controlled video sets, consisting of 9 object-control examples and 10 camera-control examples. Each question presented one input image, its corresponding motion control, and four generated videos from SG-I2V, MOFT, GWTF, and \MODELNAME. The model names were hidden, and the order of the four videos was randomized for each question. A total of 20 participants completed the anonymous survey.

Participants rated each generated video on a 1--5 Likert scale, where 1, 3, and 5 indicate very poor, moderate, and very good performance, respectively. We evaluated three independent criteria: motion alignment, physical plausibility, and visual quality. Motion alignment measures whether the generated video follows the user-defined motion control. For object control, this focuses on whether the target object moves toward the specified direction or location; for camera control, this focuses on whether the scene-level change matches the specified camera motion. Since the control signal is intentionally coarse, participants were instructed to judge motion alignment mainly by the overall displacement or trajectory rather than fine-grained object articulation. Physical plausibility measures whether the generated motion appears natural and consistent with common physical expectations, such as object-scene interactions and camera movement. Visual quality measures the overall fidelity of the generated video, including artifacts, shape distortions, blurriness, and flickering. Participants were instructed to evaluate these three criteria independently.

\begin{table}[t]
\centering
\caption{Human evaluation results on 19 video sets, including 9 object-control and 10 camera-control examples. Scores are reported as mean~$\pm$~std on a 1--5 Likert scale.}
\label{tab:human_evaluation}
\setlength{\tabcolsep}{7.6pt}
\begin{tabular}{lcccc}
\toprule
Metric & SG-I2V & MOFT & GWTF & \MODELNAME \\
\midrule
Motion alignment $\uparrow$ 
& $2.78 \pm 1.47$ 
& $2.13 \pm 1.30$ 
& $3.39 \pm 1.51$ 
& $\mathbf{4.35 \pm 0.95}$ \\
Physical plausibility $\uparrow$ 
& $3.67 \pm 1.23$ 
& $3.62 \pm 1.19$ 
& $1.77 \pm 1.06$ 
& $\mathbf{4.01 \pm 0.91}$ \\
Visual quality $\uparrow$ 
& $3.56 \pm 1.25$ 
& $3.69 \pm 1.11$ 
& $2.33 \pm 1.42$ 
& $\mathbf{3.78 \pm 1.07}$ \\
\bottomrule
\end{tabular}
\end{table}

As shown in Table~\ref{tab:human_evaluation}, \MODELNAME obtains the highest score across all three criteria. In particular, \MODELNAME achieves the best motion alignment score, indicating that participants found its generated videos to follow the specified controls more faithfully than the baselines. This supports our quantitative results based on controllability-oriented metrics such as FTD.

The lower scores of SG-I2V and MOFT are mainly due to their limited controllability and backbone quality. In many examples, they either remain nearly static or fail to follow the specified motion. This issue is especially noticeable in camera control, where they often do not align well with the intended 3D viewpoint change or scene-level camera motion. In addition, since both methods are built on SVD-based backbones, their outputs generally show lower visual fidelity compared with recent DiT-based models.

The comparison with GWTF is also informative. Although GWTF is a finetuned model with the same mask- and optical-flow-based control interface, it receives substantially lower scores in physical plausibility and visual quality. This is particularly pronounced in camera-control examples, where the evaluation is conducted on real-scene videos and GWTF often produces severely degraded or unnatural results. This supports our observation that finetuning for motion control can reduce generalization to real-world scenes. Moreover, GWTF sometimes follows imperfect or synthetic warps too rigidly, which can amplify unnatural controls and lead to implausible motion or distorted content. In contrast, \MODELNAME applies motion guidance through early query-warped attention while preserving the pretrained DiT prior during subsequent denoising. This allows the model to convert coarse user-provided controls into more plausible scene-level motion rather than directly overfitting to the warp signal.

These trends can be directly observed in the supplementary videos used for the human evaluation. Overall, the results suggest that training-free control can provide competitive controllability while better preserving perceptual realism.

% \begin{figure*}[ht]
%     \centering
%     \includegraphics[width=\linewidth]{supp/figs/vipseg_labeling.pdf}
%     \caption{Examples of our annotated samples from the VIPSeg dataset. A polygon mask is manually drawn around the target object in the first frame and warped along the indicated yellow motion vectors.}
%     % \hhk{VIPSeg sample ID는 숫자만으로 unique하게 특정할 수 없음. 특히 154로 시작하는 케이스가 2개여서 그냥 확실하게 전체 case id 써주는 것이 어떤지. (왼:1112_Ld4O3NhZ5-U , 오:154_6-pyJNd1h3I)}
%     \label{fig:vipseg_labeling}
% \end{figure*}

% \begin{figure*}
%     \centering
%     \includegraphics[width=\linewidth]{supp/figs/camera_gui.pdf}
%     \caption{
%     Our GUI for camera motion control. We implement an interactive interface that uses an external depth estimator to define camera motion from a single input image and to output the optical flow fed to the model. Users can press the \texttt{i} and \texttt{o} keys to zoom the virtual camera in and out, and draw a path with the mouse to move the camera across the image plane in arbitrary directions.
%     % Our GUI for camera motion control. 우리는 외부 depth estimator를 활용해서 입력 이미지에 대한 카메라 움직임을 정의해서 모델에 넣을 optical flow를 출력하는 GUI를 구현했다. keyboard 입력 i와 o로 각각 카메라 줌인, out을 할 수 있고, 마우스 클릭으로 경로를 그려 카메라를 이미지 plane? 위에서 사방으로 이동시킬 수 있다.
%     }
%     \label{fig:camera_gui}
% \end{figure*}

\section{Datasets}
\label{sec:datasets}

To evaluate \textit{local object motion control} and \textit{camera motion control}, we use the VIPSeg \cite{vipseg} and DL3DV \cite{dl3dv} datasets, respectively.
For reliable evaluation, we curate subsets of videos in which the target motion is clearly observable. The selected video IDs are included in our codebase and will be released publicly.
Below, we describe how each dataset is curated, preprocessed, and labeled for our experiments.
% 우리는 local object motion control과 camera motion control을 각각 평가하기 위해 각각 VIPSeg dataset과 DL3DV 데이터셋을 사용했다. 우리는 평가의 유효성을 높이기 위해서 타겟하는 모션이 잘 드러나는 비디오를 선별하였으며, 선별한 video의 ID는 우리 code에 넣어놨고 public하게 공개할 예정이다. 아래는 각각의 데이터셋을 어떻게 선별하고, 전처리하고, 라벨링했는지에 대한 자세한 명세이다. 
% \textcolor{red}{VIDEO ID 꼭 txt 나 그런 형태로 해서 코드에 포함시키기!!!!}
% \ym{The video IDs for each dataset we selected are publicly available in our codebase.}

\begin{figure*}[t]
    \centering
    \includegraphics[width=\linewidth]{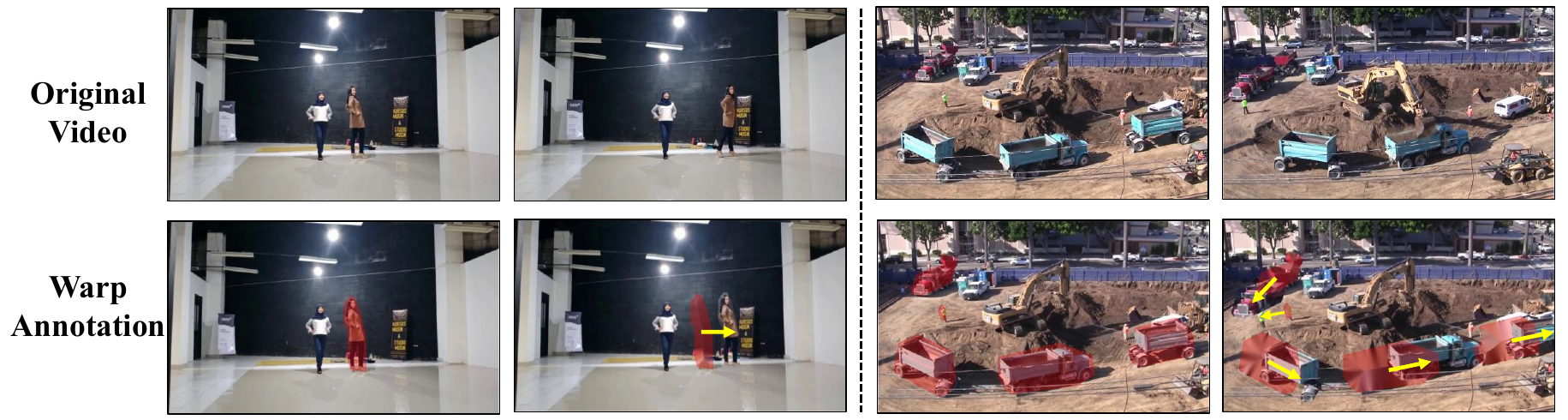}
    \caption{Examples of our annotated samples from the VIPSeg dataset. A polygon mask is manually drawn around the target object in the first frame and warped along the indicated yellow motion vectors.}
    % \hhk{VIPSeg sample ID는 숫자만으로 unique하게 특정할 수 없음. 특히 154로 시작하는 케이스가 2개여서 그냥 확실하게 전체 case id 써주는 것이 어떤지. (왼:1112_Ld4O3NhZ5-U , 오:154_6-pyJNd1h3I)}
    \label{fig:vipseg_labeling}
\end{figure*}

\myparagraph{VIPSeg.}
VIPSeg \cite{vipseg} is a video panoptic segmentation dataset providing 3,536 videos and 84,750 frames with pixel-level annotations. 100 cases that primarily exhibited local object motion rather than camera movement were selected from the validation set of VIPSeg.
Since \MODELNAME and GWTF \cite{gowiththeflow} require mask-based warping as input, we reconstructed the trajectories into the required format by annotating them with GWTF’s labeling tool, which yielded a transformation matrix capturing the frame-to-frame motion. Fig.~\ref{fig:vipseg_labeling} illustrates several examples of our annotated VIPSeg samples. For models such as SG-I2V \cite{sgi2v} and MOFT \cite{moft} that instead require bounding boxes and corresponding trajectories as inputs, we drew the smallest axis-aligned rectangle that fully enclosed the object and defined its center as the initial trajectory point. We then applied the transformation matrix to this center point to acquire its position in subsequent frames. These transformed trajectory points, together with the corresponding bounding boxes, were used as the motion prompts.

\myparagraph{DL3DV.}
DL3DV \cite{dl3dv} is a scene-level video dataset captured by moving cameras (handheld or drone) around real-world environments, so most videos are dominated by camera motion. We randomly sample 100 videos from the DL3DV 1K subset and resize and center-crop them to a resolution of $480\times 832$. Because the original frame rate is high and does not guarantee sufficient motion within a short clip, we subsample every 4 frames and keep 49 frames per video.
We use RAFT \cite{raft} to extract optical flow, which serves as the warping input for both \MODELNAME and GWTF \cite{gowiththeflow}. For methods that require camera motion to be specified via bounding boxes and trajectories, such as SG-I2V~\cite{sgi2v} and MOFT~\cite{moft}, we randomly place four $60\times 60$ bounding boxes and use CoTracker2~\cite{cotracker} to obtain the trajectories of their centers, which are then provided as model inputs.

\begin{figure*}[t]
    \centering
    \includegraphics[width=\linewidth]{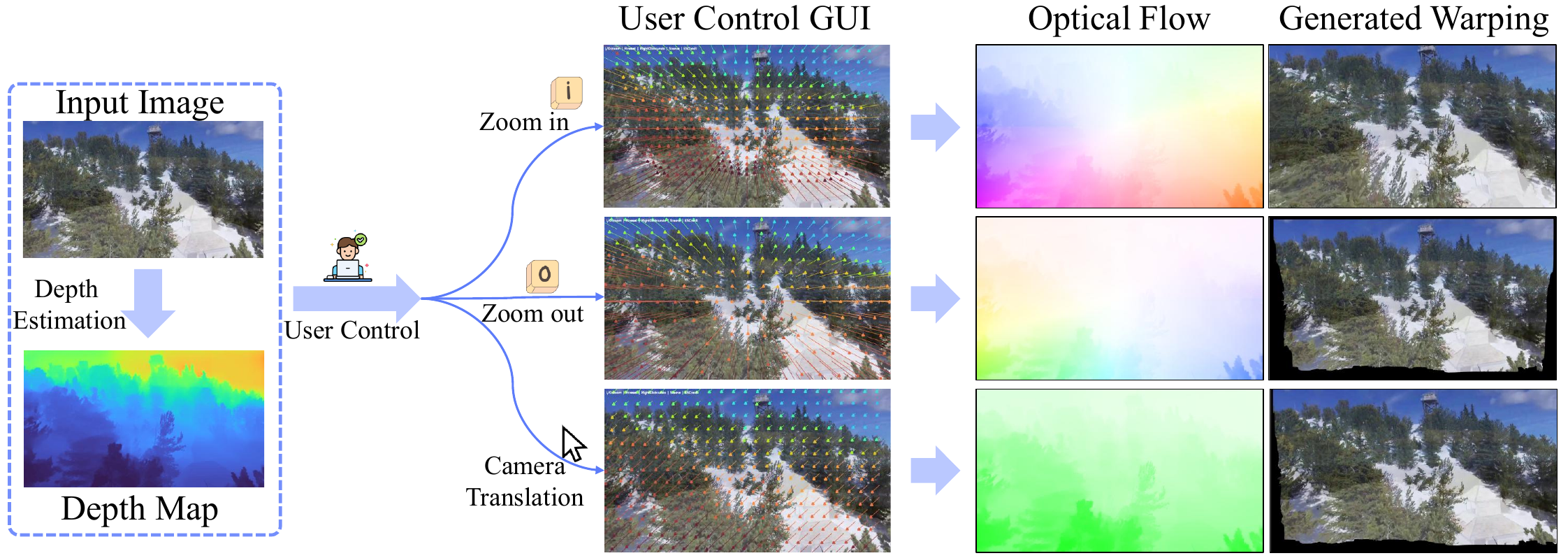}
    \caption{
    Our GUI for camera motion control. We implement an interactive interface that uses an external depth estimator to define camera motion from a single input image and to output the optical flow fed to the model. Users can press the \texttt{i} and \texttt{o} keys to zoom the virtual camera in and out, and draw a path with the mouse to move the camera across the image plane in arbitrary directions.
    % Our GUI for camera motion control. 우리는 외부 depth estimator를 활용해서 입력 이미지에 대한 카메라 움직임을 정의해서 모델에 넣을 optical flow를 출력하는 GUI를 구현했다. keyboard 입력 i와 o로 각각 카메라 줌인, out을 할 수 있고, 마우스 클릭으로 경로를 그려 카메라를 이미지 plane? 위에서 사방으로 이동시킬 수 있다.
    }
    \label{fig:camera_gui}
\end{figure*}

% \myparagraph{Interfaces for User-Defined Warping.} 
\myparagraph{Interfaces for User-Defined Warping.} 
To incorporate user-defined motion into the video generation process, we developed labeling tools. For object-level control, we directly adopted the labeling tool provided by GWTF~\cite{gowiththeflow}, enabling efficient annotation of object-centric, trajectory-based motion. In contrast, for camera-level control, we designed a new labeling tool tailored to our experimental settings. We leveraged the depth estimator~\cite{depthpro} used in GWTF to obtain an image's depth map, based on which we implemented zoom in/out and camera translation operations. This depth-based control mechanism provides a simple way to model camera motion and allows users to specify camera control through optical flow. A brief overview of the camera-control GUI is shown in Fig.~\ref{fig:camera_gui}.
% 사용자가 정의한 모션을 비디오 생성 과정에 반영하기 위해 전용 레이블링 도구를 구축하였다. Object control의 경우, GWTF에서 제공하는 레이블링 도구를 그대로 활용하여 객체 중심의 궤적 기반 모션을 효율적으로 주석화할 수 있도록 하였다. 반면, camera control을 위해서는 우리의 실험 환경에 맞춘 별도의 레이블링 도구를 새롭게 구현하였다.

% 정교한 camera control을 실현하기 위해, 우리는 GWTF에서 사용한 depth estimator 모델을 활용하여 각 프레임의 깊이 정보를 추정한 뒤, 이를 기반으로 zoom in/out 및 camera translation을 가능하게 하는 제어 시스템을 설계하였다. 이러한 depth 기반 제어 방식을 통해 다양한 카메라 움직임을 정밀하게 반영할 수 있으며, 최종적으로는 optical flow를 활용하여 사용자가 보다 직관적이고 간편하게 camera control을 수행할 수 있도록 하였다.

\begin{figure*}[t]
    \centering
    \includegraphics[width=\linewidth]{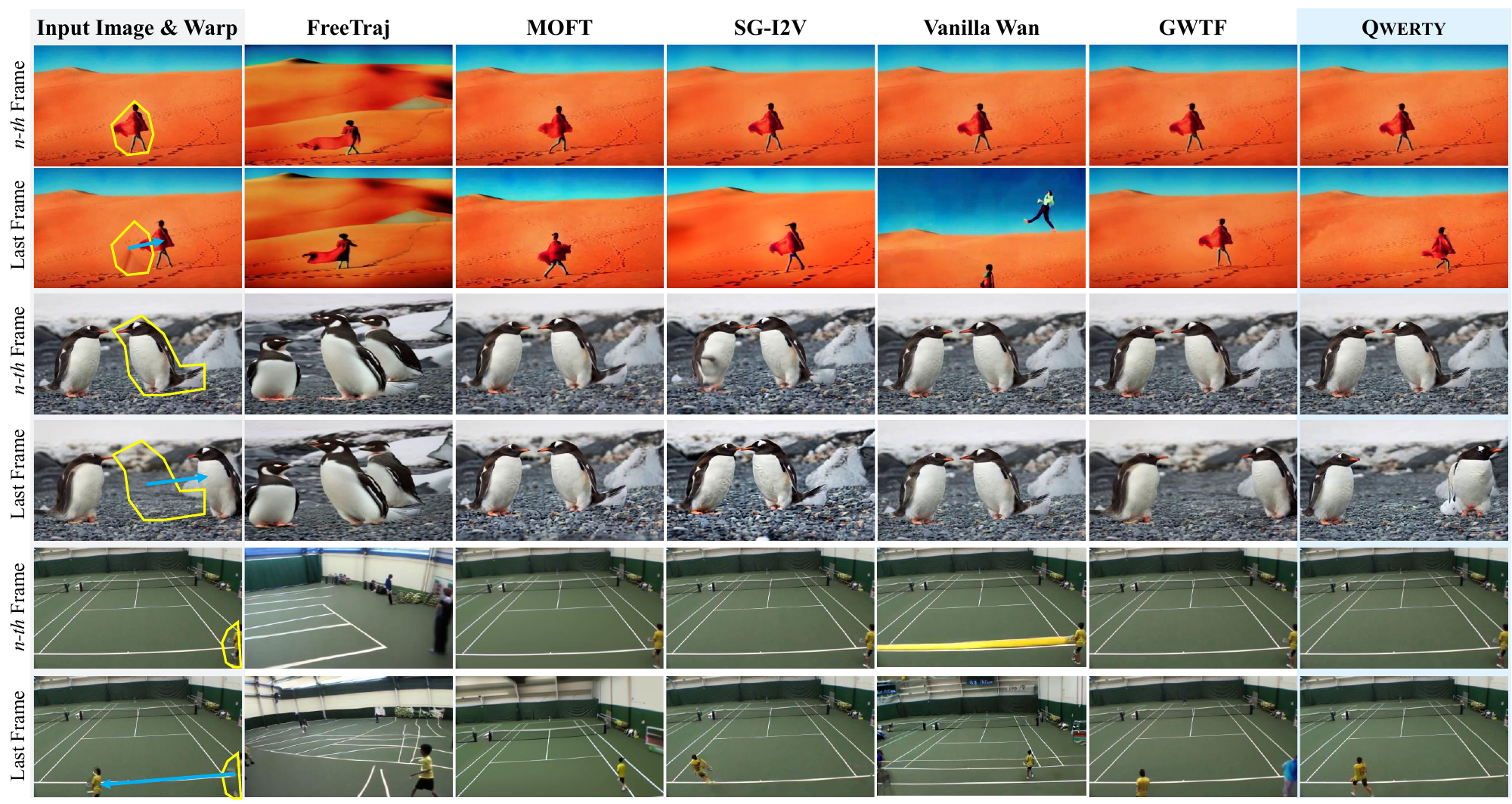}
    \caption{
    Qualitative comparison of object motion control methods. The yellow polygon indicates the region of the object to be controlled, while the blue arrow indicates motion guidance via the user-specified warp.
    Baseline models generally show little to no movement, fail to follow the control signal, produce degraded videos (FreeTraj), introduce artificial objects (Vanilla Wan, GWTF), or follow the motion in an unnatural manner (GWTF). 
    In contrast, \MODELNAME faithfully follows the user-guided motion while producing contextually coherent and artifact-free object movement.
    % 베이스라인 모델들은 전반적으로 움직임이 거의 없거나, control을 따르지 않거나, 비디오가 degrade되거나 (FreeTraj), 인공물을 소환하거나 (Vanilla Wan, GWTF), 부자연스럽게 움직임만 따라간다 (GWTF). 반면 \MODELNAME은 user 입력에 따른 움직임을 잘 따라가면서도 아티팩트 없이 맥락에 맞는 자연스러움 움직임을 만들어낸다.
    }
    \label{fig:comp_obj}
\end{figure*}

% \begin{figure*}[t]
%     \centering
%     \includegraphics[width=\linewidth]{supp/figs/comparison_object.pdf}
%     \caption{
%     Qualitative comparison of object motion control methods. The yellow polygon indicates the region of the object to be controlled, while the blue arrow indicates motion guidance via the user-specified warp.
%     Baseline models generally show little to no movement, fail to follow the control signal, produce degraded videos (FreeTraj), introduce artificial objects (Vanilla Wan, GWTF), or follow the motion in an unnatural manner (GWTF). 
%     In contrast, \MODELNAME faithfully follows the user-guided motion while producing contextually coherent and artifact-free object movement.
%     % 베이스라인 모델들은 전반적으로 움직임이 거의 없거나, control을 따르지 않거나, 비디오가 degrade되거나 (FreeTraj), 인공물을 소환하거나 (Vanilla Wan, GWTF), 부자연스럽게 움직임만 따라간다 (GWTF). 반면 \MODELNAME은 user 입력에 따른 움직임을 잘 따라가면서도 아티팩트 없이 맥락에 맞는 자연스러움 움직임을 만들어낸다.
%     }
%     \label{fig:comp_obj}
% \end{figure*}

\section{Additional Qualitative Results}
In this section, we provide additional motion-controlled video generation results obtained with \MODELNAME. We encourage readers to watch the accompanying \texttt{.mp4} videos included in the supplementary material, as static frames alone cannot fully convey the temporal coherence and realism of the generated motion.
% In this section, we provide additional motion-controlled video generation results obtained with \MODELNAME. We \textbf{\textcolor{red}{strongly encourage}} readers to watch the accompanying \texttt{.mp4} videos included in the supplementary material, as static frames alone cannot fully convey the temporal coherence and realism of the generated motion.
% 이 섹션에서는 \MODELNAME의 추가적인 motion controlled video generation 결과를 제공한다. 우리가 supplementary material 제출물에 .mp4 파일로 첨부한 결과 비디오들을 시청하는 것을 *강하게 권장*한다.

\begin{figure*}[ht]
    \centering
    \includegraphics[width=\linewidth]{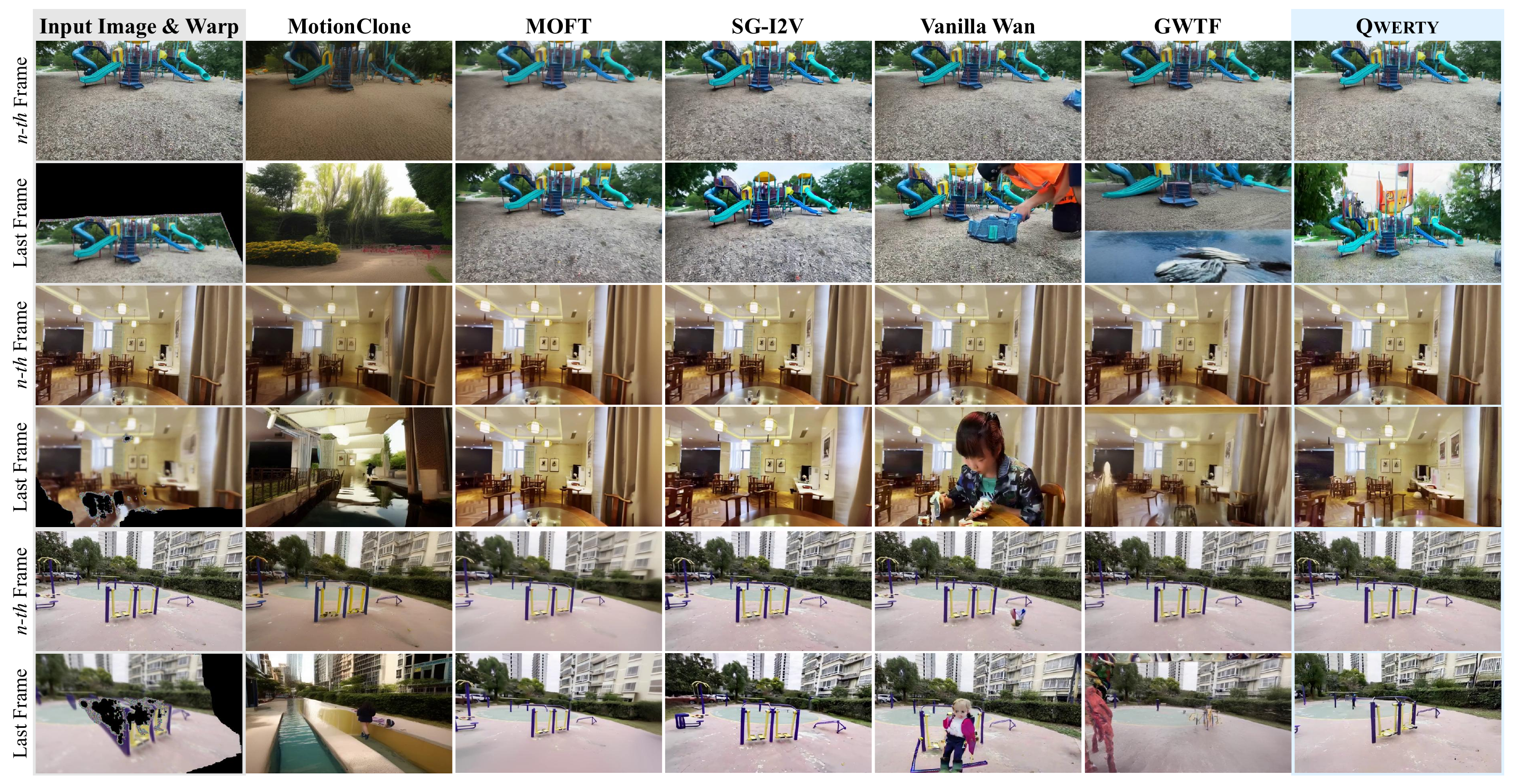}
    \caption{
    Qualitative comparison of camera motion control methods. \MODELNAME not only produces videos that closely follow the input optical flow, but also naturally fills out-of-view regions and warping-induced holes using the pretrained model’s generative capability. 
    In contrast, other methods either fail to align the camera motion precisely with the optical flow (MOFT, SG-I2V), struggle to maintain scene consistency (MotionClone), or inadvertently introduce artificial objects into the scene (Vanilla Wan, GWTF).
    % Qualitative comparison of camera motion control methods. \MODELNAME은 입력 optical flow에 잘 align된 비디오를 만들어낼 뿐만 아니라, 화면 밖 영역과 hole 부분을 pretrained 모델의 생성능력으로 자연스럽게 잘 채워넣는다. 반면에 다른 모델들은 카메라 움직임이 optical flow와 정밀하게 align되지 않거나 (MOFT, SG-I2V), 화면의 일관성을 유지하지 못하거나 (MotionClone), 인공물을 소환(Vanilla Wan, GWTF)한다.
    }
    \label{fig:comp_cam}
\end{figure*}

\begin{figure*}[ht]
    \centering
    \includegraphics[width=\linewidth]{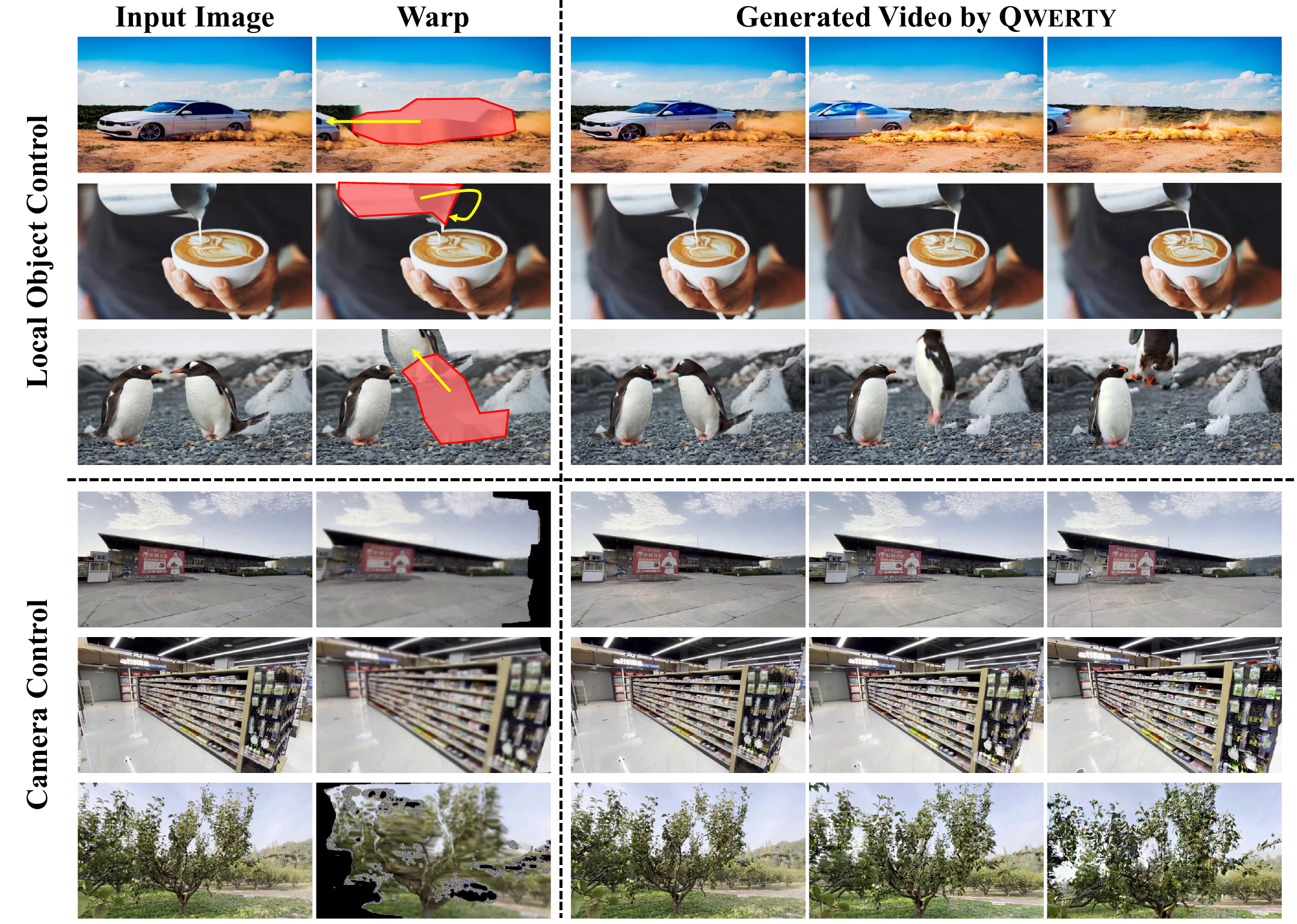}
    \caption{
    More motion control results of \MODELNAME. The red polygon indicates the region of the object to be controlled, while the yellow arrow indicates motion guidance via the user-specified warp. \MODELNAME successfully handles challenging cases, such as making a penguin jump higher or controlling camera motion while preserving fine details in the leaves.
    % 우리는 펭귄이 높이 뛰어오르는 장면이나, 나뭇잎의 디테일을 유지하면서 카메라 움직임을 컨트롤해야하는 하드한 컨트롤 케이스도 잘 수행한다.
    }
    \label{fig:addtional_qual}
\end{figure*}

\myparagraph{More Results of \MODELNAME.}
Figs.~\ref{fig:comp_obj} and~\ref{fig:comp_cam} compare our method with the baselines on object motion control and camera motion control, respectively. \MODELNAME closely follows the user-specified motion guidance while producing visually natural motion. In contrast, the strongest baseline, GWTF, is finetuned to strictly follow the user-provided warp. As a result, it often adheres to the given trajectory very rigidly, but produces object motions that are inconsistent with the scene context and severely degraded videos under real-scene camera motion. These results highlight the importance of training-free approaches, which better preserve the pretrained generative capabilities of the model. These differences become especially clear when viewing the attached video files. Additional qualitative examples generated by \MODELNAME are provided in Fig.~\ref{fig:addtional_qual}.
% Fig.~\ref{}와 Fig.~\ref{}에서는 각각 object motion control과 camera motion control에 대한 결과를 baseline과 비교하여 보여주고 있다. 우리의 \MODELNAME은 사용자의 motion guide를 잘 따르면서도 가장 자연스러운 모션을 만든다. 가장 강력한 비교군인 GWTF는 user가 제공하는 warp를 따르도록 finetuning된 모델이라 제공된 warping을 매우 엄격하게 따르지만, 맥락과 맞지않는 부자연스러운 object 모션을 만들고 real scene에 대한 camera motion에 대해서는 매우 망가진 비디오를 생성하는 것을 볼 수 있다. 이는 우리가 별도로 첨부한 비디오 파일을 보면 명확하게 볼 수 있다. 추가적으로, Fig.~\ref{fig:addtional_qual}에서 \MODELNAME의 더 많은 생성 결과들을 확인할 수 있다.

\begin{figure*}[ht]
    \centering
    \includegraphics[width=\textwidth]{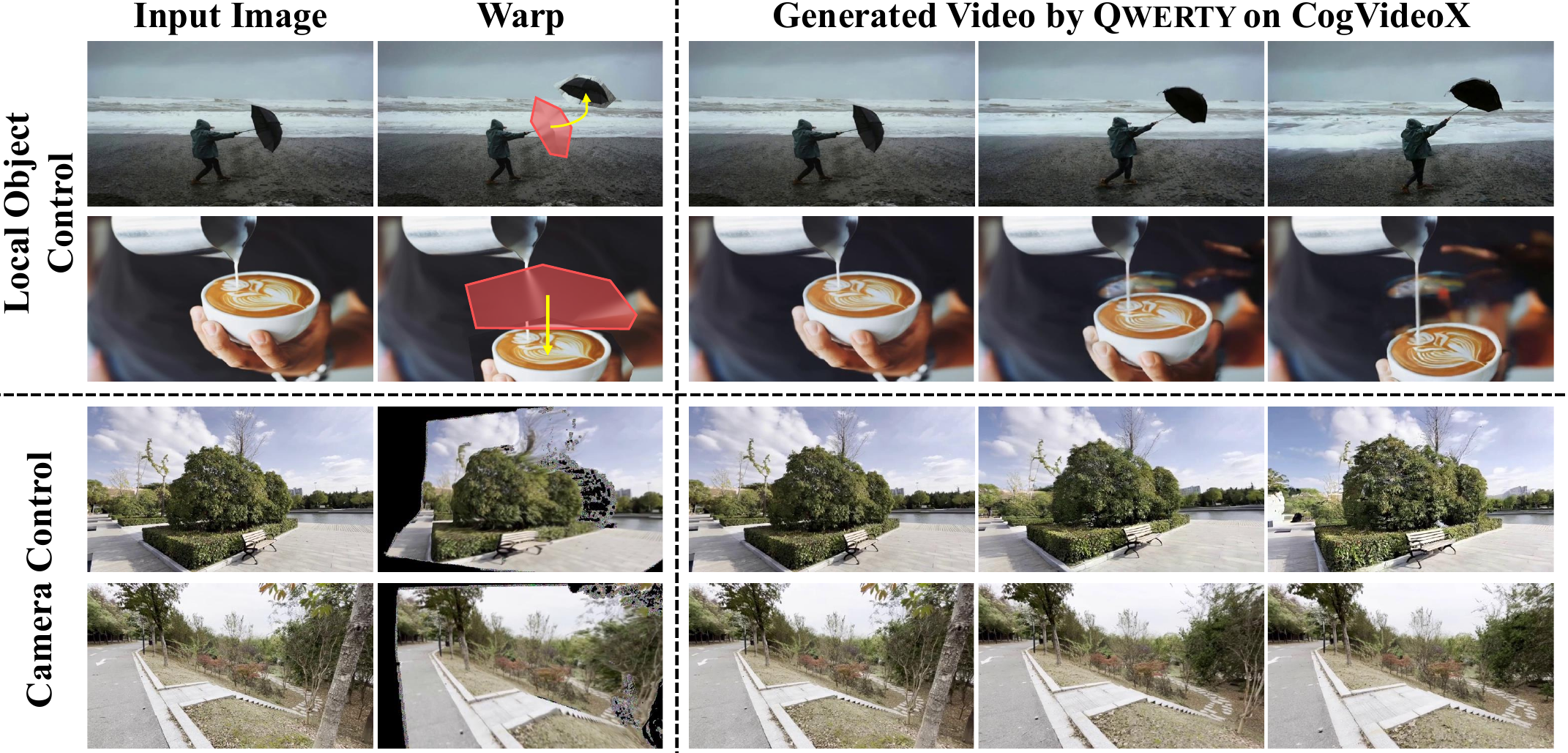}
    \caption{
    Qualitative results of \MODELNAME on the CogVideoX backbone. Similar to the observations on Wan, \MODELNAME effectively controls both object and camera motion on CogVideoX in a training-free manner, showing similar behavior across the two DiT backbones and suggesting potential applicability to other DiT-based models.
    }
    \label{fig:cogvideox}
\end{figure*}

% \input{supp/tables/cogvideox_table}
\myparagraph{Results on CogVideoX.}
We also report results of applying \MODELNAME on CogVideoX instead of Wan. As shown in Fig.~\ref{fig:cogvideox}, our method successfully controls both object motion and camera motion while maintaining high video quality on CogVideoX. This suggests that the proposed approach may also be applicable across different DiT-based I2V backbones with standard 3D attention.
% We also report results of applying \MODELNAME on CogVideoX instead of Wan. As shown in Fig.~\ref{fig:cogvideox}, our method successfully controls both object motion and camera motion while maintaining high video quality on CogVideoX. This demonstrates that the proposed approach can be applied in a fairly model-agnostic manner across different DiT-based I2V backbones.
% \myparagraph{Results on CogVideoX.}
% 우리는 \MODELNAME을 Wan이 아닌 CogVideoX 위에서 구현한 결과를 제공한다. Fig.~\ref{}는 object motion 및 camera motion control을 잘 되고 높은 퀄리티의 비디오를 만든다는 것을 보여준다. 따라서, 우리의 방법은 다양한 DiT 모델에 범용적으로 적용될 수 있음을 보여준다.

% \myparagraph{Failure cases.}

% \begin{figure*}[t]
%     \centering
%     \animategraphics[loop, controls, width=\columnwidth]{12}{supp/frames/D_DL3DV/}{0}{48}
%     \caption{Can you identify the end in the above video? \textit{Best viewed with Acrobat Reader. Click the video to play the animation clips}. We also give these examples in the \textbf{supplementary video}.}
% \end{figure*}

\section{Discussion}
In this section, we provide a detailed discussion of the experimental findings and insights derived from our study.
% 우리는 본 연구를 통해 얻은 실험결과에 대한 분석과 인사이트를 아래에서 자세히 discussion한다.

\subsection{Validity of Evaluation Metrics}
\label{discuss:evaluation}
Through our experiments, we observed that several commonly used evaluation metrics are not well aligned with the goals of user-guided motion control. In particular, FID and VBench scores tend to favor models that produce little to no motion. When a video contains significant object or background movement, or when new regions that were not present in the first frame appear in later frames, the generated video is naturally disadvantaged in VBench dimensions such as subject/background consistency, motion smoothness, and temporal flickering. Moreover, newly revealed regions must be synthesized entirely by the model. These regions often differ from the early-frame distribution in ways that are penalized by FID, which compares frames at the image level without considering motion direction or temporal context.

In practice, as shown in Table~\ref{tab:main} of the main paper, background-centric videos from DL3DV with minimal motion, such as those produced by Vanilla Wan (no motion control) or our unsuccessful Wan-based SG-I2V, can even obtain FID and VBench scores that are comparable to or higher than those of methods like GWTF or \MODELNAME. On the other hand, on object-centric scenes such as VIPSeg, where these models do exhibit noticeable motion, their FID and VBench scores become worse. This indicates that such metrics are meaningful only when comparing models that produce motion of similar magnitude and dynamics. 
By comparison, metrics that explicitly account for temporal coherence and motion fidelity, such as FVD and FTD, are more robust to these factors. These metrics better capture both perceptual video quality and motion alignment, and in our experiments, their rankings were more consistent with the qualitative outcomes observed in the supplementary videos. This observation is further supported by the human evaluation in Sec.~\ref{sec:human_eval}, where \MODELNAME is preferred across motion alignment, physical plausibility, and visual quality. These results suggest that motion-aware metrics and perceptual judgments better reflect the trade-off between control fidelity and visual realism than any single appearance-oriented metric.

% 우리는 user-guide 기반으로 motion control을 하는 본 태스크에서 몇몇 evaluation metric이 의도한 성능에 대해 모델을 줄세우는 데에 적합하지 않다고 느꼈다. FID와 VBench score는 특히 camera motion control에서 아무 움직임을 보이지 않는 경향이 큰 Vanilla Wan (본문의 Table 1)이나 Wan 위에서 구현된 SG-I2V (Table~\ref{tab:sgi2v_dit})에서 더 좋았다. 움직임으로 인해 object 또는 background가 크게 이동하고, 첫 프레임에 없던 새로운 영역들이 화면에 등장하게 되면, 정지해 있는 비디오에 비해 VBench(subject/background consistency, motion smoothness, temporal flikering) 측면에서 불리할 수 밖에 없다. 특히, 첫 프레임에 없는 영역이 비디오에 등장하게 되면, 그 부분은 모델이 자신의 생성 능력으로 만들어내야 하므로, 모션의 방향성 등을 감안하지 않고 단순히 이미지 단위에서만 두 비디오를 비교하는 FID 측면에서 특히 불리하다. 이는 Vanilla Wan이나 Wan 기반 SG-I2V가 움직임을 보이는 object-centric한 비디오에서는 GWTF나 \MODELNAME보다 FID와 VBench가 안좋은데, 움직임이 잘 없는 배경-centric한 비디오에서는 상대적으로 이 스코어들이 더 뛰어나지는 결과를 초래한다. 따라서 이러한 메트릭들은 동등한 수준의 motion dynamic을 보이는 모델들 사이에서 비디오 퀄리티를 평가하는 데에서만 유의미하다고 할 수 있다. 이에 비해 비디오 자체의 외적 유사성과 모션 유사성을 평가하는 FVD와 FTD는 이러한 변수에 대해서 더욱 강건하게 모델을 평가할 수 있고, 실제로 우리가 실제 눈으로 확인한 성능과 일치하는 결과를 보여주었다.

\subsection{Limitations}
\label{sec:limitations}
\begin{figure*}[t]
    \centering
    \includegraphics[width=\linewidth]{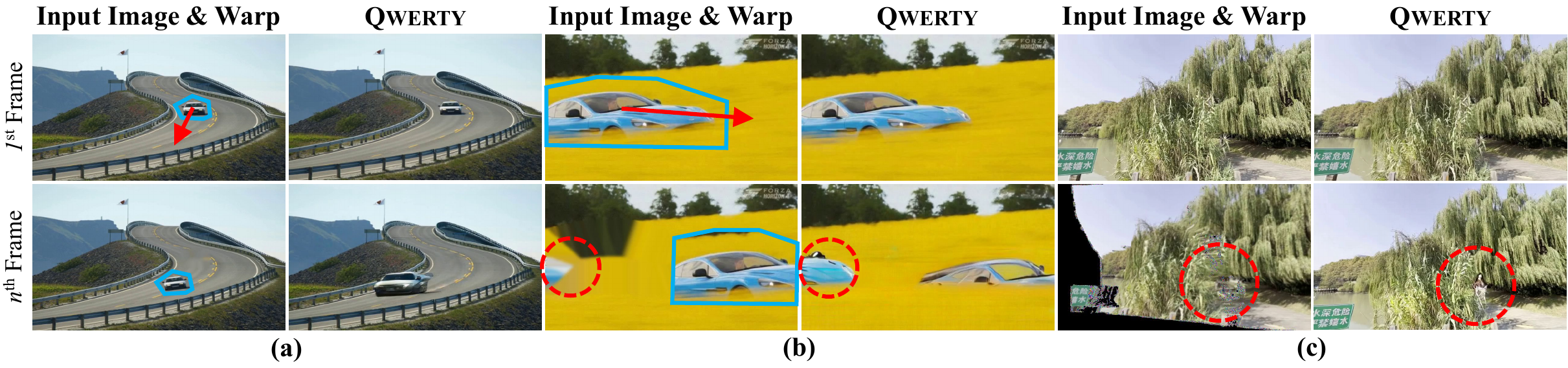}
    \caption{
    Failure cases of \MODELNAME. A detailed analysis of these cases is provided in Sec.~\ref{sec:limitations}.
    }
    \label{fig:failure_case}
\end{figure*}

We explore several failure cases in Fig.~\ref{fig:failure_case} and analyze the technical limitations of our approach.

First, because the variational autoencoder (VAE) in DiT compresses the number of latent frames (typically by a factor of four), the motion control signal can only be applied at temporally sparse positions. As a result, the generated video may occasionally fail to be perfectly aligned in time with the user-defined motion. In Fig.~\ref{fig:failure_case}(a), for example, the descending car moves farther within a single frame than specified by the given warping. We also observe slight jittering in the generated motion during camera motion control. We conjecture that this arises from the temporally sparse injection of the motion signal, where the optical flow is accumulated over four frames to match the temporal compression factor. This issue does not appear in object motion control, where the transformation from frame 1 to frame $n$ is explicitly defined, unlike optical flow defined only between consecutive frames, as described in Sec.~\ref{sec:imp_details_ours}. This reflects a fundamental disadvantage of video DiTs that operate on temporally compressed latents compared to U-Net-based models that do not, particularly in the context of training-free motion control. Nevertheless, since DiTs generally achieve superior video quality and richer motion dynamics, developing training-free motion control methods that preserve the generation capacity of pretrained DiTs still remains an important direction.

Second, in object motion control, copy artifacts can occur if the user does not fully cover the entire object with the mask while specifying a warping that moves the whole object. Fig.~\ref{fig:failure_case}(b) shows a case where the object touches the frame boundary, making it difficult for the user to mask all of its pixels. Under our hole-filling strategy (Sec.~\ref{sec:hole_filling}), the remaining hole may then be partially filled with object tokens, causing the object to be duplicated in that region (red dashed circle). This is a failure mode that can be mitigated by careful mask specification on the user side.

% \begin{figure*}[t]
%     \centering
%     \includegraphics[width=\linewidth]{supp/figs/fig_failure.pdf}
%     \caption{
%     Failure cases of \MODELNAME. A detailed analysis of these cases is provided in Sec.~\ref{sec:limitations}.
%     }
%     \label{fig:failure_case}
% \end{figure*}

Third, in camera control, the model may sometimes generate new, unintended objects while filling the holes created by camera motion. In Fig.~\ref{fig:failure_case}(c), the camera movement reveals a region behind a tree that was not visible in the first frame, resulting in a hole (red dashed circle). Our model fills this hole by synthesizing a person in that region. Similar object \emph{summoning} behavior has also been observed in GWTF, a finetuned DiT-based model. Although such unintended objects may be undesirable in some cases, they also reflect the model’s ability to create novel content, which can be seen as a double-edged sword. Since DiT-based I2V models can also take a text prompt as input, one can combine text-based control with our method at generation time to either suppress or encourage such phenomena according to the user’s intent.

Finally, backbone artifacts can be amplified under strong motion control. 
In our Wan-based camera-control experiments, artifact-like structures from the pretrained backbone can become more visible when user-defined motion imposes large geometric changes. 
Since camera control displaces or reprojects the entire scene, such amplified backbone artifacts may appear as wobble artifacts.
Thus, these failures reflect the interaction between large motion guidance and the fixed pretrained backbone, rather than artifacts introduced solely by query warping.
% 우리는 Fig.~\ref{fig:failure_case}에서 failure case들을 탐색하고 우리의 기술적인 limitation을 분석한다.
% first, DiT의 VAE가 latent 프레임 수를 압축(보통 4배)하기 때문에 temporal하게 sparse한 motion control signal이 적용될 수밖에 없다. 이로 인해 가끔 user가 정의한 motion과 시간적으로 완전히 align되지 않은 비디오를 생성할 수 있다. Fig.~\ref{fig:failure_case}(a)를 보면 내려오는 자동차가 주어진 warping에 비해 같은 프레임에서 더 많이 갔다. 이는 시간축으로 압축된 latent를 사용하는 비디오 DiT가 그렇지 않은 U-Net 기반 모델보다 training-free motion control에서 근본적으로 불리한 부분이라고 할 수 있다. 그러나, DiT가 전반적인 비디오 퀄리티나 모션 dynamic 등 성능이 훨씬 좋기 때문에, pretrained DiT의 generation capacity를 그대로 유지하면서 motion control이 가능한 training-free 접근 방식은 여전히 중요하다.
% second, object motion control 시 오브젝트의 전체 부분을 mask로 제대로 선택하지 않고 오브젝트 전체를 움직이는 warping을 주면 복사 아티팩트가 생길 수 있다. Fig.~\ref{fig:failure_case})(b)는 오브젝트가 프레임 경계에 붙어 있어서 유저가 완벽히 object의 모든 픽셀을 마스킹하지 못한 경우이다. 이렇게 되면 우리의 hole-filling 전략 (Sec.~\ref{fig:hole_filling})에 따라 hole이 일부 object 토큰으로 채워지게 되어서 그 부분에 object가 복제될 수 있다 (빨간 점선 원). 이는 사용자가 주의하면 방지할 수 있는 실패 케이스이다.
% 마지막, camera control 시 생긴 hole을 모델이 스스로 채우면서 가끔 새로운 object를 생성해낼 수가 있다. Fig.~\ref{fig:failure_case})(c)를 보면 카메라 움직임으로 인해 첫 프레임에는 보이지 않던 나무 뒤 공간이 드러나 hole이 되게 된다 (빨간 점선 원). 우리 모델은 이 hole 영역에서 사람을 소환시킨 것을 볼 수 있다. 이렇게 object를 소환하는 현상은 DiT 기반의 finetuning 모델인 GWTF에서도 종종 관찰되었다. 모델의 생성 능력으로 인해 사용자가 원하지 않는 object가 나타날 수도 있지만, 그만큼 더욱 novel한 비디오를 생성할 수 있다는 것이므로 이는 양날의 검이다. DiT 기반 I2V 모델은 text prompt를 함께 받을 수 있으므로, 생성 시 text 기반의 컨트롤을 병행해서 이러한 현상을 의도대로 억제하거나 유도할 수 있다.

\subsection{Future Works}
This work suggests several promising directions for future research. 

First, achieving temporally fine-grained training-free motion control directly in the temporally compressed latent space of video DiTs remains a key challenge, as noted in the first limitation discussed in Sec.~\ref{sec:limitations}. One possible direction is to better understand the temporal compression mechanism of the VAE and design motion-guidance strategies that are consistent with this behavior. Another is to explicitly involve the VAE in the motion-control loop, so that even though the latent sequence is temporally downsampled, the resulting motion signal can still remain temporally precise.

Second, it is important to explore more computationally efficient training-free motion control techniques tailored to DiT architectures. Our current method still relies on latent optimization, which significantly increases sampling time. Developing alternatives that preserve a comparable level of controllability while reducing or partially amortizing the optimization overhead would make DiT-based motion control more practical in real-world applications, and we believe that our query-warping–only variant can serve as a useful stepping stone in this direction.

Finally, there is a need for evaluation metrics specifically tailored to motion control that jointly capture both appearance fidelity and motion similarity, as discussed in Sec.~\ref{discuss:evaluation}. Although FVD partially serves this role, it measures the similarity between distributions of videos and therefore does not provide a fine-grained, per-sample assessment for each guided sample. Designing metrics that directly quantify how well a generated video follows a given motion specification, while still accounting for overall visual quality, is an important direction for future work.

\end{document}